\documentclass[lettersize,journal]{IEEEtran}
\usepackage{amsmath,amsfonts}
\usepackage{algorithmic}
\usepackage{algorithm}
\usepackage{array}
\usepackage[caption=false,font=normalsize,labelfont=sf,textfont=sf]{subfig}
\usepackage{textcomp}
\usepackage{stfloats}
\usepackage{url}
\usepackage{verbatim}
\usepackage{graphicx}
\usepackage{cite}
\usepackage{color}
\usepackage{multirow}
\usepackage{booktabs}
\usepackage{bbding}
\usepackage{threeparttable}
\usepackage{makecell}

\begin{document}

\title{EventDiff: A Unified and Efficient Diffusion Model Framework for Event-based Video Frame Interpolation}

\author{Hanle Zheng, Xujie Han, Zegang Peng, Shangbin Zhang, Guangxun Du, Zhuo Zou,~\IEEEmembership{Senior Member,~IEEE}, \\Xilin Wang, Jibin Wu,~\IEEEmembership{Member,~IEEE}, Hao Guo, Lei Deng,~\IEEEmembership{Senior Member,~IEEE}
\thanks{This work was supported in part by the National Natural Science Foundation of China (No. 62276151 and 62411560155), and Chinese Institute for Brain Research, Beijing. (Corresponding authors: Hao Guo and Lei Deng)}
\thanks{Hanle Zheng, Zegang Peng, and Lei Deng are with the Center for Brain Inspired Computing Research (CBICR), Department of Precision Instrument, Tsinghua University, Beijing, China (e-mail: zhl22@mails.tsinghua.edu.cn; pzg24@mails.tsinghua.edu.cn; leideng@mail.tsinghua.edu.cn).}
\thanks{Xujie Han and Hao Guo are with the College of Computer Science and Technology, Taiyuan University of Technology, Jinzhong, China (e-mail: xjhan@link.tyut.edu.cn; guohao@tyut.edu.cn).}
\thanks{Shangbin Zhang and Guangxun Du are with The Information Science Academy of China Electronics Technology Group Corporation, Beijing, China. (e-mail: zsbss1989@163.com; duguangxun@126.com)}
\thanks{Zhuo Zou is with the State Key Laboratory of Integrated Chips and Systems, School of Information Science and Technology, Fudan University, Shanghai, China (e-mail: zhuo@fudan.edu.cn).}
\thanks{Xilin Wang is with the Engineering Laboratory of Power Equipment Reliability in Complicated Coastal Environments, Tsinghua Shenzhen International Graduate School, Tsinghua University, Shenzhen, China (e-mail: wang.xilin@sz.tsinghua.edu.cn).}
\thanks{Jibin Wu is with the Department of Data Science and Artificial Intelligence and the Department of Computing, The Hong Kong Polytechnic University, Hong Kong SAR (e-mail: jibin.wu@polyu.edu.hk).}
}



\maketitle

\begin{abstract}
Video Frame Interpolation (VFI) is a fundamental yet challenging task in computer vision, particularly under conditions involving large motion, occlusion, and lighting variation. Recent advancements in event cameras have opened up new opportunities for addressing these challenges. While existing event-based VFI methods have succeeded in recovering large and complex motions by leveraging handcrafted intermediate representations such as optical flow, these designs often compromise high-fidelity image reconstruction under subtle motion scenarios due to their reliance on explicit motion modeling. Meanwhile, diffusion models provide a promising alternative for VFI by reconstructing frames through a denoising process, eliminating the need for explicit motion estimation or warping operations. In this work, we propose EventDiff, a unified and efficient event-based diffusion model framework for VFI. EventDiff features a novel Event-Frame Hybrid AutoEncoder (HAE) equipped with a lightweight Spatial-Temporal Cross Attention (STCA) module that effectively fuses dynamic event streams with static frames. Unlike previous event-based VFI methods, EventDiff performs interpolation directly in the latent space via a denoising diffusion process, making it more robust across diverse and challenging VFI scenarios. Through a two-stage training strategy that first pretrains the HAE and then jointly optimizes it with the diffusion model, our method achieves state-of-the-art performance across multiple synthetic and real-world event VFI datasets. The proposed method outperforms existing state-of-the-art event-based VFI methods by up to 1.98dB in PSNR on Vimeo90K-Triplet and shows superior performance in SNU-FILM tasks with multiple difficulty levels. Compared to the emerging diffusion-based VFI approach, our method achieves up to 5.72dB PSNR gain on Vimeo90K-Triplet and 4.24$\times$ faster inference. Furthermore, EventDiff demonstrates strong extensibility by achieving competitive results on event-based motion deblurring, showcasing its great potential as a unified framework for event-enhanced visual generation tasks.
\end{abstract}


\section{Introduction}
\IEEEPARstart{V}{ideo} rame interpolation (VFI) is a challenging task in the field of computer vision, aiming to restore high-quality, high-frame-rate videos from low-frame-rate video inputs. This area has gained widespread attention due to its various applications, such as slow-motion generation, video compression, and video frame recovery. Traditionally, frame-based VFI methods rely on linear or quadratic motion assumptions and approximations of motion fields. These methods \cite{hu2022many,huang2022real,jiang2018super,jin2023enhanced,park2020bmbc,park2021asymmetric,shen2024ladder} usually utilize optical flow estimation to compensate for motion between ground-truth frames and warp intermediate frames. There are also some flow-free methods \cite{niklaus2017video,niklaus2021revisiting} that directly synthesize interpolated frames without estimating the motion field. However, frame-based VFI methods have inherent limitations when handling complex scenes with non-linear motion, object occlusions and brightness variations, which significantly impact their performance.

The development of bio-inspired dynamic vision sensors (DVS) \cite{amir2017low} has created new opportunities in computer vision tasks. By detecting brightness changes in each pixel, the DVS, also known as the event camera, generates sparse event streams asynchronously. offering advantages such as low latency, high dynamic range and lightweight data representation. Recently, many studies \cite{tulyakov2021time,tulyakov2022time,kim2023event,ma2024timelens,liu2024video} have leveraged event data to enhance performance in VFI tasks. Similar to traditional frame-based methods, these approaches can be categorized into two types: motion-based methods, which use event information for more accurate non-linear motion estimation, and synthesis-based methods, which directly fuse dynamic event features and static frame features in a flow-free and end-to-end learning framework. However, both of these methods still require careful design of intermediate representations such as optical flow and grayscale images to guide the fusion and warping process. Although these designs can guide the feature extraction process better, such handcrafted components become performance bottlenecks and limit the generalization capability. For example, although the state-of-the-art (SOTA) event-based methods \cite{ma2024timelens} perform well in high-motion scenes, their performance deteriorates under subtle motion variations. To date, no event-based VFI method consistently delivers robust performance across a wide variety of scenes.

Diffusion Models (DM) \cite{song2020denoising,song2020score} have emerged as a powerful generative paradigm, showing impressive capabilities in visual synthesis tasks, such as text/audio/image-conditioned visual generation \cite{ho2022video,singer2022make}, visual completion \cite{yin2023nuwa} and visual editing tasks \cite{xing2024dynamicrafter}. The VFI task could also benefit from the application of diffusion models for frame generation \cite{voleti2022mcvd,danier2024ldmvfi,lyu2024frame}, which can be viewed as a form of image-conditioned video generation. Diffusion models offer a promising alternative to handcrafted intermediate supervision by directly learning the reverse diffusion process through progressive corruption of ground-truth data with Gaussian noise during training, showing great potential to improve VFI performance. As the first effort that targets VFI using frame-based latent diffusion models (LDM), a recent work \cite{danier2024ldmvfi} proposes the LDMVFI method comprising an autoencoding model that projects images into a latent space and a denoising U-Net that performs the reverse diffusion process. However, this approach focuses more on the perceptual quality of the generated frames, often lagging in reconstruction accuracy (e.g., PSNR and SSIM). Furthermore, the inference speed of LDMVFI is significantly slower compared to other frame-based VFI methods, primarily due to the large number of iteration denoising steps required. As a result, existing DM frameworks are not yet suitable for VFI tasks.

\begin{figure}[!t]
\centering
\includegraphics[width=1.0\columnwidth]{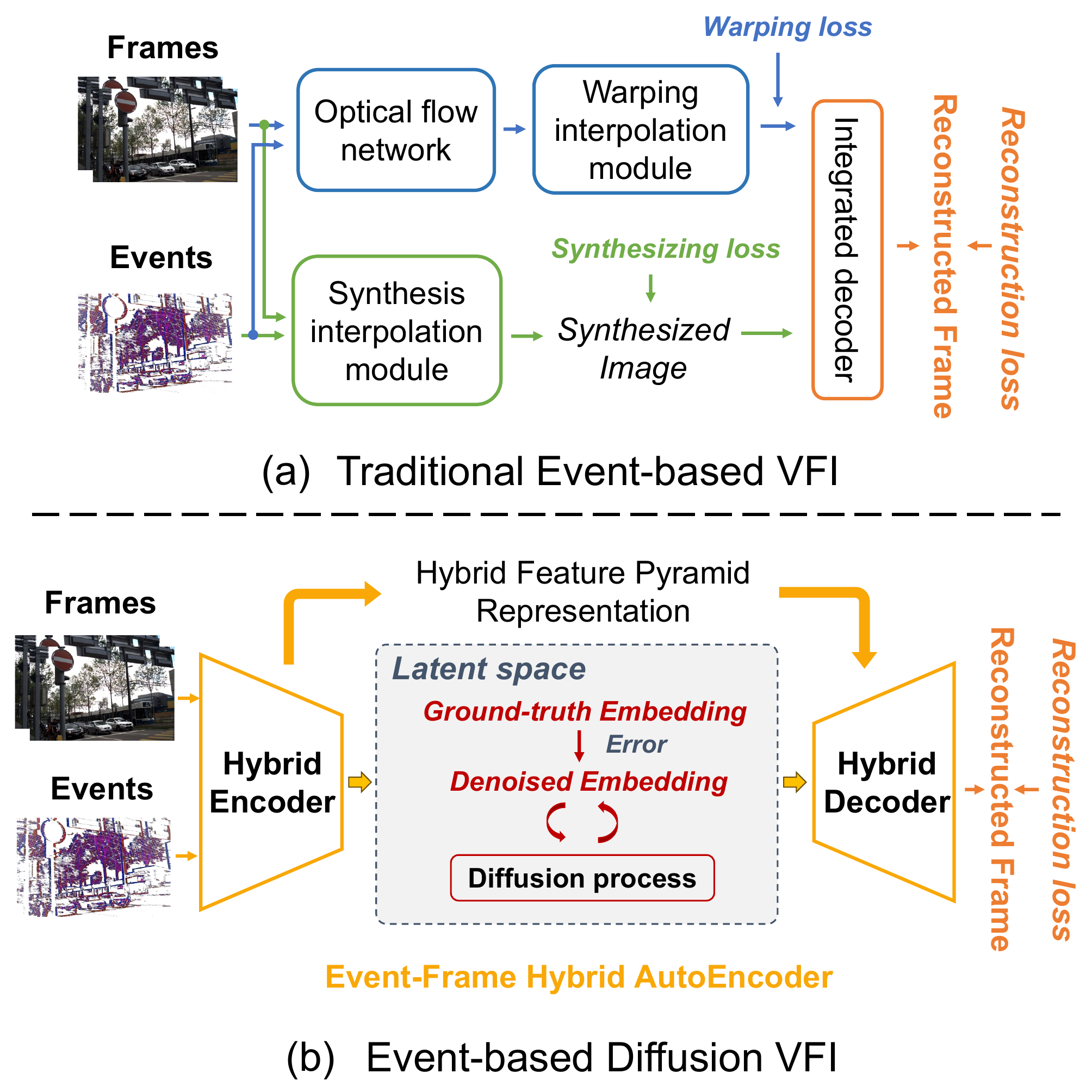}
\caption{\textbf{EventDiff compared with previous Event-based VFI methods.} (a) The traditional event-based VFI methods, such as Timelens \cite{tulyakov2021time} et al., use multi-component architectures with separate warping, synthesis and refinement modules. These methods rely on intermediate products like optical flow serving as manually designed auxiliary elements for fusing frames and events, which make bottlenecks and lead to suboptimal results. (b) The proposed EventDiff employs a unified and efficient autoencoder framework that directly optimizes against ground-truth frames without requiring handcrafted intermediate products. Instead, it utilizes a diffusion model to reconstruct the ground-truth embedding in the latent space, enabling more accurate VFI preformance. }
\label{fig1}
\end{figure}

To address above challenges in frame-based, event-based and diffusion-based VFI methods, we propose \textbf{EventDiff}, a unified and efficient diffusion model framework for event-based VFI. It is straightforward, powerful and computationally efficient. First, we construct an Event-Frame Hybrid AutoEncoder (HAE), which incorporates a lightweight Spatial-Temporal Cross Attention (STCA) module. This module enables effective fusion of dynamic event streams and static frames, capturing rich spatial-temporal information. The HAE encodes the boundary frame-event pairs into a hybrid feature pyramid representation, while embedding the ground-truth frame into the latent space, serving as a target for reconstruction during the diffusion process. Next, we perform the diffusion process in the latent space to reconstruct the ground-truth embedding, conditioned on the hybrid features from neighboring frame-event pairs. Similar to the latent diffusion model, EventDiff is trained in two stages. In the first stage, the HAE is pretrained under the guidance of ground-truth inputs, which generates the ground-truth embedding and conditional features, laying the foundation for the subsequent diffusion process. Simultaneously, it learns a hybrid feature pyramid representation for the decoding process. In the second stage, we reconstruct the ground-truth embedding through the diffusion process and jointly optimize the pretrained HAE to minimize potential performance degradation.

Unlike other event-based VFI methods (as illustrated in Fig \ref{fig1}), EventDiff eliminates the need for complex sub-network designs, such as warping/synthesis modules and handcrafted intermediate loss functions. Instead, it performs end-to-end global optimization optimization through a unified autoencoder architecture and generate results by the diffusion model, making it more robust across diverse and challenging VFI scenarios. Compared to other diffusion-based VFI methods, the introduction of event data significantly improves performance by incorporating additional spatial-temporal information. The joint optimization further reduces the number of inference steps in the diffusion process, significantly improving inference speed. 

In short, our contributions can be summarized as follows:
\begin{itemize}
    \item We propose EventDiff, a unified and efficient diffusion model framework for event-based VFI. Unlike other event-based VFI methods, it performs global optimization through a unified autoencoder for hybrid modalities and executes the diffusion process in the latent space, offering a straightforward, powerful and computationally efficient VFI solution. Furthermore, EventDiff can be extended to motion deblurring tasks, highlighting its strong potential as a general framework for event-enhanced visual generation.
    \item We design a general Event-Frame Hybrid AutoEncoder (HAE) architecture, which introduces the Spatial-Temporal Cross Attention (STCA) mechanism to seamlessly integrate event and frame modalities, enabling effective hybrid feature extraction.
    \item A joint optimization algorithm is proposed to minimize the error between the denoised embedding and ground-truth embedding in the latent space during the diffusion process, while simultaneously reducing the reconstruction error of the frame from the denoised embedding. This results in improved performance and more efficient frame interpolation
    \item Extensive experiments on synthetic and real-world VFI benchmarks evidence that EventDiff outperforms existing frame-based, event-based, and diffusion-based VFI methods, achieving up to 1.98dB PSNR gain over the best event-based method and 5.72dB PSNR gain over emerging diffusion-based methods on Vimeo90K-Triplet, with up to 4.24$\times$ faster inference.
\end{itemize}

\section{Related Works}

\subsection{Frame-based Video Frame Interpolation}

Frame-based VFI has made significant strides, with methods generally falling into two main categories: flow-based and kernel-based approaches. Flow-based methods \cite{hu2022many,huang2022real,jiang2018super,jin2023enhanced,park2020bmbc,park2021asymmetric,shen2024ladder} utilize optical flow to represent the motion between frames, which is a widely adopted technique for motion representation. In these approaches, intermediate frames are synthesized by estimating optical flow, either with or without pretrained flow models. On the other hand, kernel-based methods implicitly model motion using kernels. These methods \cite{niklaus2017video,niklaus2021revisiting} generate intermediate frames by applying local convolution operations to boundary frames. A key advantage of kernel-based approaches is the ability to capture complex local motion patterns. Recent advancements in deformable convolution kernels have further enhanced the flexibility and performance of these methods \cite{cheng2020video,cheng2021multiple,chen2021pdwn}. Beyond these two primary categories, some other approaches attempt to combine flow- and kernel-based methods \cite{bao2019memc,danier2022st}. More recently, the rise of transformer-based models has led to the incorporation of transformer architectures to better capture long-range dependencies and effectively model large and non-rigid motions \cite{lu2022video,zhang2023extracting,li2023amt,plack2023frame}. However, frame-based VFI methods face inherent challenges when dealing with complex scenes involving non-linear and large motion, object occlusions and variations in brightness, which would substantially degrade their performance.

\begin{figure*}[!htbp]
\centering
\includegraphics[width=\textwidth]{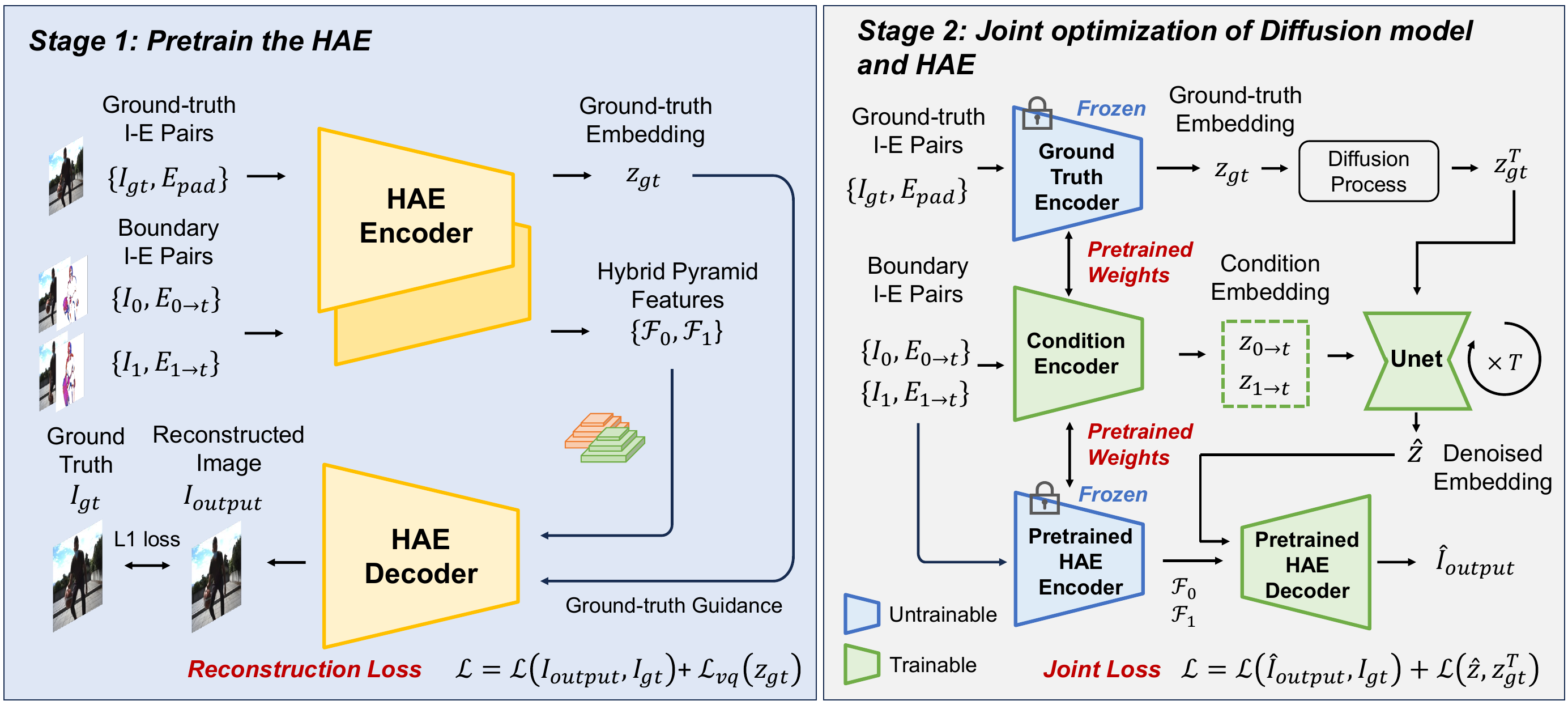}
\caption{\textbf{Illustration of the EventDiff framework.}  The framework consists of two training stages. In Stage 1 (left), we train the Event-Frame Hybrid AutoEncoder (HAE) using ground-truth inputs, enabling it to capture hybrid pyramid features and project the ground-truth into the latent space. In Stage 2 (right), the ground-truth inputs are removed and a diffusion model is employed to reconstruct the ground-truth embedding within the latent space.}
\label{framework}
\end{figure*}

\subsection{Event-based Video Frame Interpolation}

The DVS \cite{amir2017low}, i.e., the event camera, offers significant advantages over traditional frame cameras. With its low response latency ($ \leq $ 1us), wide dynamic range ($ \geq $120 dB) and energy efficiency, it is well-suited for capturing rapid scene changes, such as fast-moving objects, large motions and sudden brightness variations. Thus, the VFI task can benefit from event data, as it provides supplementary information regarding the changes occurring between frames. One type of event-based VFI methods uses the event data for more accurate motion estimation. These methods, such as Timelens \cite{tulyakov2021time} and others \cite{tulyakov2022time,ma2024timelens}, combine motion-based warping with synthesis techniques. Among them, Kim et al. \cite{kim2023event} achieve advanced performance by supplementing both image-level and event-level motion fields and incorporating an advanced transformer decoder. Ma et.al \cite{ma2024timelens} reach the SOTA results when dealing with large motion. Another approach focuses on flow-free methods, directly synthesizing interpolated frames without motion estimation, as seen in EFI-Net \cite{paikin2021efi} and Superfast\cite{gao2022superfast}. A recent study \cite{liu2024video} proposes a novel method that directly synthesizes interpolated frames based on inter-frame references reconstructed from event data. However, existing event-based VFI methods still require careful design of intermediate products for fusing frames and events, along with complex sub-networks for warping or synthesis. These methods lack a unified framework capable of directly fusing both modalities and performing global optimization.

\subsection{Diffusion-based Video Frame Interpolation}

Denoising diffusion models \cite{song2020denoising,rombach2022high} have demonstrated exceptional performance in image and video generation. Recently, some studies have begun exploring their application in VFI. The MCVD model \cite{voleti2022mcvd} is the first attempt in this area. It performs a denoising diffusion process on the entire image, conditioned on past and/or future frames. However, it is designed for low-resolution video synthesis tasks and does not introduce VFI-specific innovations. Subsequently, LDMVFI \cite{danier2024ldmvfi} makes a significant advancement by projecting video frames into a latent space using an autoencoder, before applying the diffusion model for reconstruction within that space. While these similar models \cite{jain2024video,lyu2024frame,wang2024framer,shen2024dreammover} outperform traditional frame-based VFI methods in perceptual metrics, they struggle with similarity metrics like PSNR and SSIM, indicating limited performance in interpolation accuracy. More recently, Huang et al. \cite{huang2024motion} explore the integration of event data into diffusion models for VFI. Their approach utilizes the event information to estimate motion, which is subsequently fused into a VQ-MAGAN-like structure similar to LDMVFI. While this method further improves perceptual quality, it inherits the same shortcomings of LDMVFI in terms of low interpolation accuracy. In summary, existing diffusion-based methods, whether incorporating event data or not, have yet to demonstrate a clear superiority over traditional frame-based and event-based VFI methods.

\section{Method}

\subsection{Framework Overview}

 The objective of Event-based VFI is to reconstruct the intermediate frame $I_{gt}$ given the boundary Frames $I_0$ and $I_1$ along with the inter-frame event data $E_{0\rightarrow1}$. We first divide the event data into two parts: $E_{0\rightarrow t}$ and $E_{1\rightarrow t}$ based on the time step $t$ of the groud-truth frame $I_{gt}$ to be interpolated. In essence, the event stream $E$ is represented as a set of events $\{e_i\}$, where each event is denoted by $(x_i,y_i,p_i,t_i)$, with $x_i$ and $y_i$ representing the location of the event $e_i$, $p_i$ indicating its polarity (with a value of -1 or 1),  and $t_i$ representing the timestamp when the event occurs. Hence, $E_{0\rightarrow t}$ and $E_{1\rightarrow t}$ can be defined as
\begin{equation}
    E_{0\rightarrow t} = \{(x_i,y_i,p_i,t_i)\mid t_i<t\},
\end{equation}
\begin{equation}
    E_{1\rightarrow t} = \{(x_i,y_i,-p_i,t_i)\mid t_i>t\}.
\end{equation}
Subsequently, we obtain the boundary Image-Event (I-E) pairs: $\{I_0,E_{0\rightarrow t}\}$ and $\{I_1,E_{1\rightarrow t}\}$. The ground-truth I-E pair is padded as $\{I_{gt},E_{pad}\}$ where $E_{pad} = \emptyset$. 

The overall structure of EventDiff is illustrated in Fig. \ref{framework}, which is trained with two stages. In the first stage (the right of Fig. \ref{framework}), we pretrain the Event-Frame Hybrid AutoEncoder (HAE). The HAE encoder performs two key functions: first, it compresses the ground-truth I-E pairs and projects them into the latent space to generate the ground-truth embeddings; second, it extracts features from the boundary I-E pairs to produce hybrid pyramid features. The decoder then leverages these ground-truth embeddings, combining them with the hybrid pyramid features to reconstruct the intermediate frames. The training of HAE is globally optimized under the direct supervision of the ground-truth embeddings, eliminating the need for intermediate warping or synthesized outputs, thereby enhancing its overall reconstruction performance. In the second stage, our goal is to train a diffusion model to generate reconstructed denoised embeddings from the Boundary I-E pairs, aiming to estimate the ground-truth embeddings. We then replace the original denoised embeddings of HAE with the newly generated denoised embeddings to finetune HAE. Therefore, the training process will involve joint optimization of both the diffusion model and HAE.

\subsection{Stage \uppercase\expandafter{\romannumeral1}: Pretrain HAE}

\begin{figure*}[!t]
\centering
\includegraphics[width=\textwidth]{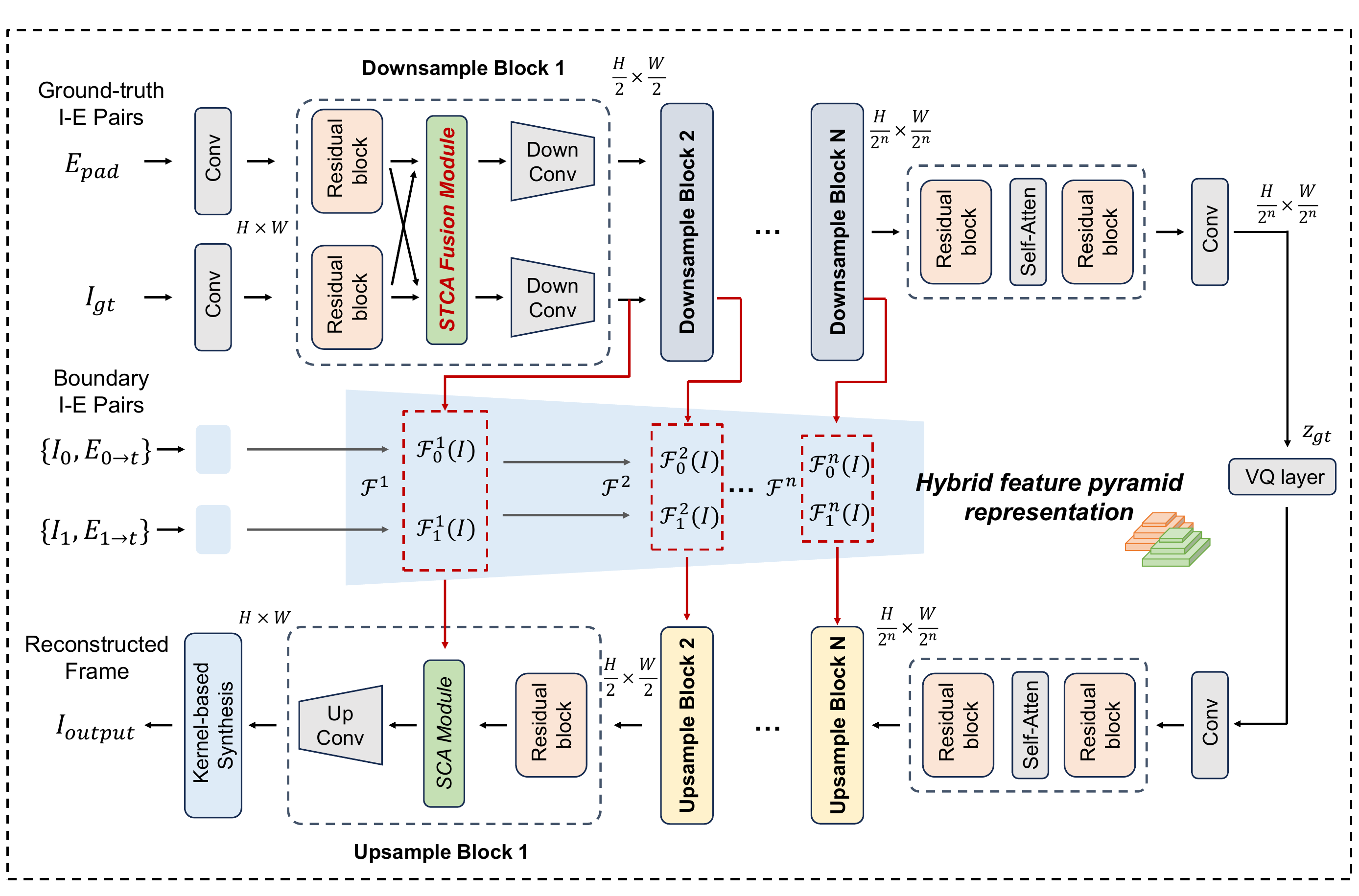}
\caption{Illustration of the Event-Frame Hybrid AutoEncoder (HAE) architecture.}
\label{HAE}
\end{figure*}

We develop the proposed HAE based on the VQFIGAN  adopted by LDMVFI \cite{danier2024ldmvfi} and VQGAN used in the latent diffusion model \cite{rombach2022high}. The HAE architecture is shown in Fig. \ref{HAE}. First, we integrate events into the standard event voxel grid as the approach used in \cite{tulyakov2021time,kim2023event} with $T$ temporal bins. For instance, the voxel representation $E^V \in \mathbb{R}^{T\times2\times H\times W}$ of the events date $E = \{(x_i,y_i,p_i,t_i)\}$ can be defined as
\begin{equation}
\begin{aligned}
E^V(m,p,x,y) &= \sum_i \delta(x_i-x)\delta(y_i-y)\delta(p_i-p)\delta(t'_i-m)\\
    t'_i &= \lfloor T\frac{t_i-t_{min}}{t_{max}-t{min}} \rfloor
\end{aligned}
\end{equation}
where $t_{min}$ and $t_{max}$ represent the start and end time of the integrated event stream $E$, and $\delta$ denotes the Dirac function. 

Next, the boundary I-E pairs are fed into the HAE encoder to generate hybrid pyramid features at different levels. The HAE encoder consists of multiple downsampling blocks, each extracting one spatial level of features. More specifically, for the images with the dimension of $H \times W$,  we extract $n$ levels of features with $n$ downsampling blocks, as defined below
\begin{equation}
\{\mathcal{F}^s_{k}(I),\mathcal{F}^s_{k}(E),s \in [1,n],k=0,1\} = \mathcal{H}(I_k,E^V_{k\rightarrow t})
\end{equation}
where $\mathcal{F}^s_{k}(I)$ and $\mathcal{F}^s_{k}(E)$ denotes the image-based and event-based feature at the level $s$. Each level of features has spatial the dimension of $\frac{H}{2^s} \times \frac{W}{2^s}$. 
 
\begin{figure}[t]
\centering
\includegraphics[width=1.0\columnwidth]{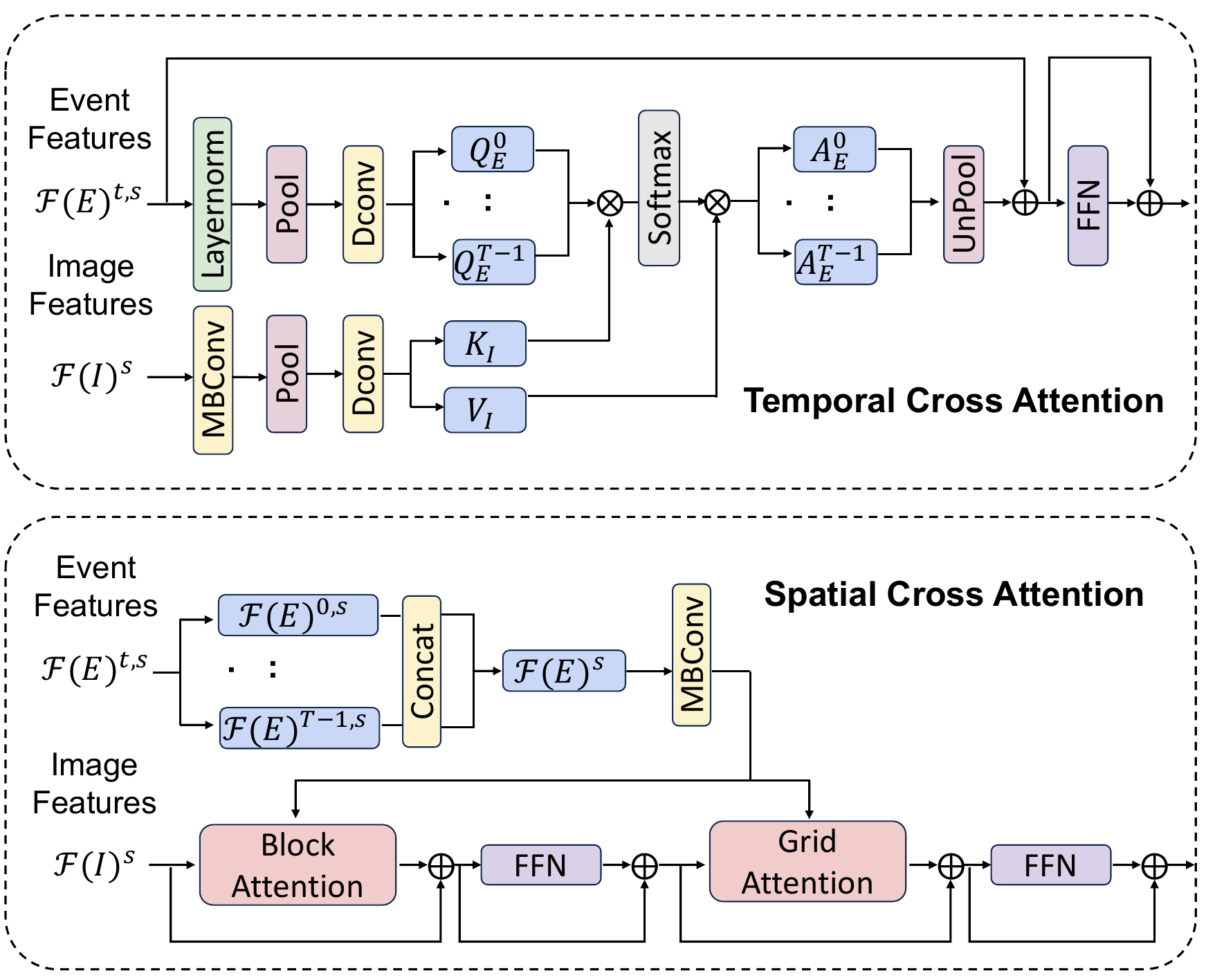}
\caption{Illustration of (a) Temporal Cross Attention (TCA) and (b) Spatial Cross Attention (SCA) modules.}
\label{stca}
\end{figure}

Each downsampling block consists of two branches, where each branch is responsible for feature extraction from one modality (either events or images). These features are fused bidirectionally between the branches in the spatio-temporal cross-attention (STCA) fusion module within the downsampling block, facilitating cross-modal feature integration. More specifically, given the input of the $l$-th downsampling block, $\mathcal{F}^{l-1}_{k}(I)$ and $\mathcal{F}^{l-1}_{k}(E)$, the output of the downsampling block, $\mathcal{F}^{l}_{k}(I)$ and $\mathcal{F}^{l}_{k}(E)$, can be defined as 
\begin{equation}
\begin{aligned}
\mathcal{O}^l_{k}(I) &= \mathcal{H}_{res}(\mathcal{F}^{l-1}_{k}(I)) \\
\mathcal{O}^l_{k}(E) &= \mathcal{H}_{res}(\mathcal{F}^{l-1}_{k}(E))  \\
\mathcal{F}^l_{k}(I) &= \mathcal{H}_{d}(\mathcal{H}_{s}(\mathcal{O}^l_{k}(I),\mathcal{O}^l_{k}(E)))\\
\mathcal{F}^l_{k}(E) &= \mathcal{H}_{d}(\mathcal{H}_{t}(\mathcal{O}^l_{k}(E),\mathcal{O}^l_{k}(I)))
\end{aligned}
\end{equation}
where $\mathcal{H}_{res}$ and $\mathcal{H}_{d}$ denote Residual block operation and downsampling convolutional operation. The STCA fusion module consists of two type of attention operation: the Spatial Cross Attention (SCA), $\mathcal{H}_{s}$, and the Temporal Cross Attention (TCA), $\mathcal{H}_{t}$, which are illustrated in Fig. \ref{stca}. 

The TCA performs local attention along the temporal dimension. Specifically, given the input event feature map $F(E) \in \mathbb{R}^{T\times\hat{C} \times \hat{H}\times \hat{W}}$ where $T,C,H,W$ respectively represents the time, channel, height, and width dimension. We first apply a $p\times p$ pooling operation to reduce its spatial resolution to $\frac{\hat{H}}{p} \times \frac{\hat{W}}{p}$. The feature map is then partitioned into $\frac{\hat{H}\hat{W}}{p^2}$ non-overlapped  sub-slices, each with a temporal sub-slice shape of $T\times \hat{C}$. For the corresponding image feature map $F(I) \in \mathbb{R}^{\hat{C'}\times \hat{H}\times \hat{W}}$ , we use the MBConv block\cite{howard2017mobilenets} to transform its channel dimension to $\hat{C}$, aligning it with the event feature map. A similar pooling and partitioning operation is applied, producing  $\frac{\hat{H}\hat{W}}{p^2}$ non-overlapping sub-slices, each with with a shape of
$1\times \hat{C}$. Cross-attention is then applied between the event and image feature maps along each temporal sub-slice, resulting
\begin{equation}
\begin{aligned}
Q_E[i,j] &= W_q F'(E)[i,j] \\
K_I[i,j] &= W_k F'(I)[i,j]  \\
V_I[i,j] &= W_v F'(I)[i,j] \\
A_E[i,j] = Softmax(Q&_E[i,j]K_I[i,j]^T / \sqrt{\hat{C}}) \cdot V_I[i,j]
\end{aligned}
\end{equation}
where $F'(E) \in \mathbb{R}^{T\times\hat{C} \times \frac{\hat{H}}{p} \times \frac{\hat{W}}{p}}$ and $F'(I)\in \mathbb{R}^{\hat{C} \times \frac{\hat{H}}{p} \times \frac{\hat{W}}{p}}$ are the transformed event and frame feature maps after pooling and MBconv operations, $W_q$, $W_k$, and $W_v$ are learnable projection matrices, and $i \in [0,\frac{\hat{H}}{p}]$, $j \in [0,\frac{\hat{W}}{p}]$ denote the spatial locations of each temporal sub-slice. The attention weights yield an attention map $\alpha[i,j] = Softmax(Q[i,j]K[i,j]^T / \sqrt{\hat{C}})$,  which quantifies the relevance of each temporal sub-slice of the event feature map in relation to the image feature. Next, the temporally aggregated feature map $A_E\in \mathbb{R}^{T\times\hat{C} \times \frac{\hat{H}}{p}\times \frac{\hat{W}}{p}}$ is reshaped back to its original spatial dimension through an unpooling operation, producing a refined feature representation that incorporates temporal contexts. In the implementation, we set the pooling scale at $p=2$ .

On the other hand, the SCA follows a similar structure to the MaxCA module proposed in the LDMVFI method. It begins by concatenating the event feature map $F(E) \in \mathbb{R}^{T\times\hat{C} \times \hat{H}\times \hat{W}}$  along the temporal dimension, reshaping it into $F'(E) \in \mathbb{R}^{T\hat{C} \times \hat{H}\times \hat{W}}$. The MBConv module is again utilized to transform the channel dimension to $\hat{C'}$, ensuring alignment with the image feature map. Subsequently, a MaxVit-like \cite{tu2022maxvit} block attention and grid attention mechanism is applied to fuse the event and image features within the spatial domain, resulting in the spatially aggregated feature map $A'_E\in \mathbb{R}^{\hat{C'} \times \hat{H}\times \hat{W}}$.

The STCA module simplifies the computationally intensive full spatial-temporal cross-attention by factorizing it into two more efficient operations: one along the spatial dimension and the other along the temporal dimension. This decomposition significantly reduces the computational cost, lowering it from $O(T\hat{H}\hat{W} \cdot\hat{H}\hat{W})$ to $O(((T/p^2+\hat{H}\hat{W})\cdot\hat{H}\hat{W})$.

In addition to the boundary I-E pairs, the ground-truth I-E pairs are also passed through a series of downsampling blocks, reducing their spatial dimensions. The output $\mathcal{F}^{n}_{gt}(I)$ are further transformed by subsequent Residual blocks and a self-attention module into the ground-truth embedding $z_{gt} \in \mathbb{R}^{{3} \times \frac{H}{2^n}\times \frac{W}{2^n}}$. During decoding, the hybrid feature pyramids $\{\mathcal{F}^s_{k}(I),s \in [1,n],k=0,1\}$ are fused with the decoded features from the quantized ground-truth embedding $z_{gt}$ via the SCA module. Finally, we generate the reconstructed frame $I_{output}$ using deformable convolution-based interpolation kernels \cite{cheng2020video,danier2024ldmvfi}.

In the first stage, the whole HAE is trained with the loss $\mathcal{L}_{stage1}$ which includes two items: the L1 loss between the reconstructed frame $I_{output}$ and the ground-truth $I_{gt}$, and the latent regularization term based on a vector quantization (VQ) layer \cite{van2017neural}. Thus, the whole loss $\mathcal{L}_{stage1}$ can be represented as
\begin{equation}
 \mathcal{L}_{stage1} =\mathcal{L}_1(I_{output},I_{gt})+\mathcal{L}_{vq}(z_{gt}).
\end{equation}

Compared to VQFIGAN, HAE introduces two key differences: (1) the inclusion of the event modality and the elaborated STCA module for more effective hybrid feature fusion; (2) the removal of the LPIPS-based perceptual loss \cite{zhang2018unreasonable} and GAN-style adversarial loss \cite{isola2017image} used in VQFIGAN while optimizing the spatial dimension of the latent space. As a result, we observe a significant improvement in interpolation accuracy during the autoencoder training stage, with much fewer parameters.

\subsection{Stage \uppercase\expandafter{\romannumeral2}: Joint optimization of  Diffusion Model and HAE}

The Stage 2 focuses on reconstructing the ground-truth embedding  based on the diffusion model. Typically, training Diffusion Probabilistic Models (DDPM) involves two key phases: a forward diffusion process and a reverse denoising process.

In our proposed method, the forward diffusion process resembles the standard latent diffusion model. The ground-truth encoder inherits the parameters of the pretrained HAE encoder in Stage 1, generating the ground-truth embedding $z_{gt}$. Random noise is then gradually added in the original embedding $z_{gt}$ over a series of $T$ time steps, increasingly producing noisy data, denoted as $z^{t'}$ for $t'=1,2,\cdots,T$. This process can be defined as a Markov chain, wherein at each time step Gaussian noise is added according to a variance schedule $\beta_t$. The forward process is modeled by
\begin{equation}
q(z^{t'}\mid z^{t'-1}) = \mathcal{N}(z^{t'};\sqrt{1-\beta_{t'}}z_{gt},\beta_{t'}I)
\label{beta}
\end{equation}
where $\mathcal{N}$ represents a Gaussian distribution, and $\beta_{t'}$ is a hyperparameter that controls the noise variance at time step $t'$. The cumulative signal-to-noise ratio is governed by the parameters ${\alpha}_{t'} =1-\beta_{t'}$ and $\bar{\alpha}_{t'} = \prod_{s=1}^{t'}\alpha_s$. This defines the relationship between $z_{gt}$ and $z^{t'}$ as
\begin{equation}
z^{t'} = \sqrt{\bar{\alpha}}z_{gt}+\sqrt{1-\bar{\alpha}}\epsilon
\end{equation}
where $\epsilon \sim \mathcal{N}(0,1)$ represents standard Gaussian noise.

The reverse denoising process aims to denoise the noisy sample $z^{T'}$ and recover the original data distribution. The model learns the conditional distribution $p_\theta(z^{t'-1}\mid z^{t'})$, parameterized by $\theta$ to reverse the forward process:
\begin{equation}
p(z^{t'-1}\mid z^{t'}) = \mathcal{N}(z^{t'-1};\mu_\theta(z^{t'},t'),\Sigma_\theta(z^{t'},t')).
\end{equation}
The mean $\mu_\theta(z^{t'},t)$ is estimated using the predicted noise  $\epsilon_\theta(z^{t'},t)$, generated by a denoising network. In traditional diffusion models, the denoising network is trained by minimizing the error between the true noise $\epsilon$ and the predicted noise $\epsilon_\theta$ at each time step, using the following loss function:
\begin{equation} 
\mathcal{L}_{DM} = \mathbb{E}_{z^{t'},t,\epsilon}[\lVert \epsilon-\epsilon_\theta(z^{t'},t)\rVert^2].
\end{equation}

However, since our diffusion process is applied to a compact ground-truth embedding, fewer iterations are needed to achieve accurate noise estimation. Inspired by DiffI2I \cite{xia2024diffi2i}, instead of estimating the noisy at each time step, our method trains the denoising network by directly minimizing the error between the finally estimated denoised embedding  $\hat{z}$ and the ground-truth embedding $z_{gt}$ as follows:
\begin{equation}
\mathcal{L}_{DM} = \mathcal{L}_1(\hat{z},z_{gt})
\end{equation} 
where $\mathcal{L}_1$ denotes the L1 loss, and $\hat{z}$ is the final denoised embedding after $T$ iterations from the noisy sample $\hat{z}^T = z^T$, following Eq. (\ref{ite}). Given the denoised sample $\hat{z}^{t'}$ at the time step $t'$,  The denoised sample at the time step $t'-1$ is then derived by
\begin{equation} \label{ite}
\hat{z}^{t'-1} = \frac{1}{\sqrt{\alpha_{t'}}}(\hat{z}^{t'-1}-\frac{1-\alpha_{t'}}{\sqrt{1-\bar{\alpha_{t'}}}}\epsilon_\theta(z^{t'},t',z_{0\rightarrow t},z_{1\rightarrow t})).
\end{equation}

\begin{figure}[!t]
\centering
\includegraphics[width=1.0\columnwidth]{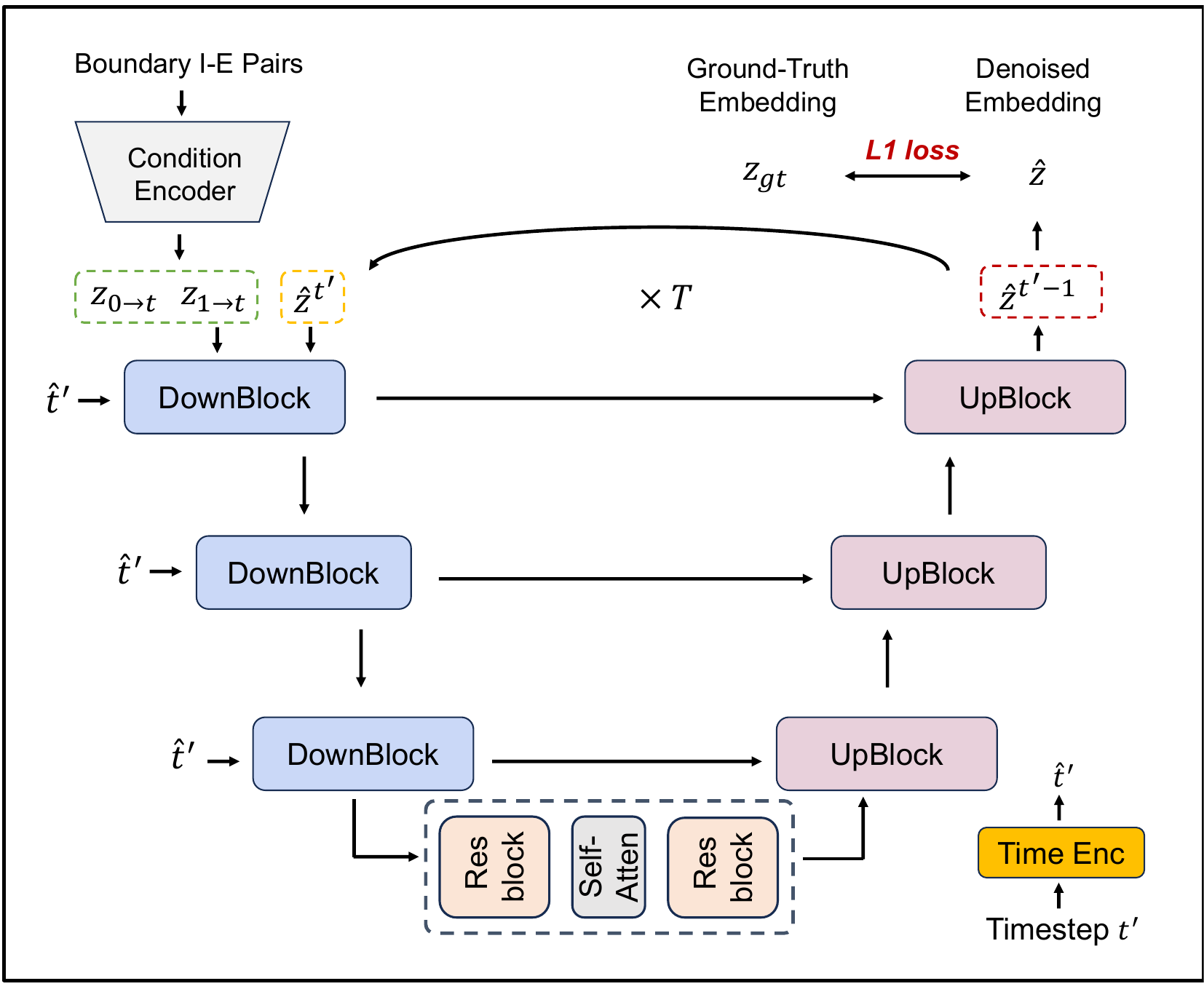}
\caption{Illustration of the U-net with a condition encoder for diffusion model.}
\label{unet}
\end{figure}

Here, $\epsilon_\theta(z^{t'},t',z_{0\rightarrow t},z_{1\rightarrow t})$ is generated by a denoising network, which is a simple U-net with a condition encoder (Fig. \ref{unet}). The condition encoder, like the ground-truth encoder, inherits the pretrained parameters from the HAE encoder in Stage 1, generating condition embeddings $z_{0\rightarrow t}$ and $z_{1\rightarrow t}$ from the boundary I-E pairs $\{I_0,E_{0\rightarrow t}\}$ and $\{I_1,E_{1\rightarrow t}\}$ respectively. The U-net  takes these condition embeddings, the diffusion sample $\hat{z}^{t'}$ and the corresponding time step to estimate $\epsilon_\theta(\hat{z}^{t'},t',z_{0\rightarrow t},z_{1\rightarrow t})$. 

Meanwhile, to alleviate the influence of the error between the estimated denoised embedding $\hat{z}$ and the ground-truth embedding $z_{gt}$, we also finetune the pretrained HAE. As a result, we perform joint optimization of the diffusion model and the HAE decoder at the same time (Fig. \ref{framework} right) in which only the ground-truth encoder that generates the ground-truth embedding template is fixed without training. The joint training loss is governed by
\begin{equation}
\mathcal{L}_{stage2} = \mathcal{L}_1(\hat{I}_{output},I_{gt})+\mathcal{L}_1(\hat{z},z_{gt})
\end{equation}
where $\hat{I}_{output}$ represents the reconstructed frame when replacing the ground-truth embedding with the estimated denoised embedding.

During the inference stage of the diffusion model, the denoising process begins with a random Gaussian noisy input and progressively estimates the denoised embedding. Our joint optimization approach significantly reduces the large number of inference iterations required in traditional diffusion models (typically 100 to 1000 steps) to only $T = 5$ steps, while maintaining high interpolation accuracy owing to the pretrained HAE from Stage 1.

\section{Experiments}

\subsection{Implementation Details}

To demonstrate the effectiveness of our proposed framework, we evaluate the methods on both synthetic and real-world event VFI datasets. 

\subsubsection{Synthetic Event VFI Datasets}

The Synthetic Event VFI Datasets includes the Vimeo90k \cite{xue2019video}, GoPro \cite{nah2017deep} and SNU-FLIM \cite{choi2020channel} datasets, commonly used to evaluate the performance of VFI methods. We first train the model on the collection of the training sets of Vimeo90k-Septuplet and Gopro, with the synthetic event data simulated using ESIM \cite{gehrig2020video}. And we further finetune the model on GoPro for better generalization on images with higher resolution. Then, we test the trained model on the testing sets of the Vimeo90k-Triplet, GoPro and SNU-FLIM.

\subsubsection{Real-world Event VFI Datasets}

We utilize the publicly available real-world event VFI datasets for evaluation: HQ-EVFI \cite{ma2024timelens}, BS-ERGB \cite{tulyakov2022time}, and High Quality Frame (HQF) \cite{stoffregen2020reducing} DAVIS240. Among these, HQ-EVFI and BS-ERGB contain both training and validation sets, with raw frames as high-resolution RGB images. While BS-ERGB offers well-aligned RGB and event data, the severe noise in some RGB images makes it less reliable for evaluation. On the other hand, HQ-EVFI provides high-quality RGB images with well-synchronized event and RGB data, making it a better real-world benchmark for Event-VFI methods. The HQF dataset, however, consists of grayscale images with a relatively low resolution of $180 \times 240$. Our process begins by finetuning the HAE model, which is pretrained on synthetic event VFI datasets, using the training set of each real-world event dataset. Then, we train the diffusion model from scratch in the second stage. Finally, we evaluate the trained model on each real-world dataset.

\subsubsection{Experimental Setting}

For training, we crop input images in Vimeo90k-Septuplet into $256 \times 256$ patches and $512 \times 512$ patches in GoPro, HQ-EVFI and BS-ERGB. During the training of Stage 1, the Adam optimizer is used with an initial learning rate of $1\times10^{-5}$, which decays to $1\times10^{-6}$ after 90 epochs. During the training of Stage 2, the parameters in U-net are optimized by the learning rate of $1\times10^{-5}$ and the parameters in pretrained HAE are optimized by the learning rate of $1\times10^{-6}$. The parameters $\beta_{t'}$ in Eq. \ref{beta} is linearly increased from $\beta_{1} = 1\times10^{-5}$ to $\beta_{T'} = 0.1$, where the number of iteration steps in the diffusion process is set to $T'=5$. All the experiments are conducted with the Pytorch framework on NVIDIA RTX 4090 GPUs.

Prior work for comparison on above datasets include three types of  methods: traditional frame-based VFI methods, diffusion-based VFI methods and event-based VFI methods. Following the setting in prior work, we use the Peak Signal to Noise Ratio (PSNR) and Structural Similarity (SSIM) to evaluate the performance of the video frame interpolation task. Moreover, since previous diffusion-based VFI methods focus on optimizing the perception quality, we further compare perceptual quality (LPIPS) with them. For frame-based and diffusion-based VFI methods, we directly interpolate intermediate frames using pretrained models to assess their performance. However, for event-based VFI methods, the references do not provide full access to all pretrained models, training codes and the dataset generation process used for evaluation. To ensure a fair and comprehensive comparison, we directly evaluate the methods assuming that the pretrained weights are available. Otherwise, we finetune several event-based VFI methods, including TimeLens, CBMNet-L, and TLXNet+ for datasets where prior performance is unavailable or pretrained weights are not released. The finetuning process is based on the implementation in \cite{ma2024timelens}. 

\begin{table*}[!htbp]
    \centering
    \caption{Quantitative performance comparison (PSNR $\uparrow$ (DB) / SSIM$\uparrow$) of results on synthetic event VFI datasets.}
    \begin{tabular}{ccccccccccc}
        \toprule
        \multirow{2}{*}{Method} & \multirow{2}{*}{Events} & \multirow{2}{*}{DM} & \multirow{2}{*}{Vimeo90K-Triplet} & \multicolumn{2}{c}{GoPro} & \multicolumn{4}{c}{SNU-FILM}  \\ \cmidrule(r){5-6} \cmidrule(r){7-10}
        &&&&7-skips&15-skips&easy&medium&hard&extreme\\
        \midrule
        ABME\cite{park2021asymmetric}  & \XSolidBrush & \XSolidBrush & 36.22/0.981&27.79/0.888&21.57/0.758&39.75/\underline{0.990}&35.84/\underline{0.979}&30.62/0.937&25.42/0.864\\
        RIFE\cite{huang2022real} & \XSolidBrush & \XSolidBrush& 35.61/0.978 &27.71/0.891&22.47/0.773&40.06/\textbf{0.991}&35.73/\textbf{0.980}&30.09/0.933&24.84/0.854\\
        IFRNet\cite{kong2022ifrnet} & \XSolidBrush & \XSolidBrush& 36.21/0.981 &27.72/0.891&22.50/0.775&40.10/\textbf{0.991}&36.12/\textbf{0.980}&30.63/0.937&25.26/0.861\\
        UNP-Net-L\cite{jin2023unified} & \XSolidBrush & \XSolidBrush& 36.42/\underline{0.982} &27.92/0.893&22.80/0.777&\textbf{40.44}/\textbf{0.991}&36.29/\textbf{0.980}&30.86/0.938&25.63/0.864\\
        VFIformer\cite{lu2022video} & \XSolidBrush & \XSolidBrush& 36.50/\underline{0.982} &27.84/0.892&22.53/0.776&\underline{40.13}/\textbf{0.991}&36.09/\textbf{0.980}&30.67/0.938&25.43/0.864\\
        EMA-L\cite{zhang2023extracting} & \XSolidBrush & \XSolidBrush& 36.64/\underline{0.982} &28.07/0.896&22.85/0.780&39.98/\textbf{0.991}&36.09/\textbf{0.980}&30.94/0.939&25.68/0.866\\
        \addlinespace
        \midrule
        MCVD\cite{voleti2022mcvd} & \XSolidBrush & \Checkmark & -&-&-&22.20/0.828&21.49/0.812&20.31/0.766&18.46/0.694\\
        LDMVFI\cite{danier2024ldmvfi} & \XSolidBrush & \Checkmark& 33.09/0.962&27.11/0.881&22.01/0.758&38.67/0.987&34.00/0.970&28.55/0.917&23.93/0.837\\
        BBD\cite{lyu2024frame} & \XSolidBrush & \Checkmark& 35.43/0.977 &27.49/0.886&22.20/0.766&39.64/0.990&34.89/0.974&29.62/0.929&24.38/0.848\\
        \addlinespace
        \midrule
        TimeLens\cite{tulyakov2021time} & \Checkmark & \XSolidBrush& 36.31/0.962 &36.93/0.982&35.18/0.974&37.50/0.977&35.69/0.967&33.88/0.957&31.78/0.941\\
        $\text{A}^2$OF\cite{wu2022video} & \Checkmark & \XSolidBrush& 36.61/0.971 &34.08/0.954&32.65/0.937&-&-&31.72/0.924&28.21/0.890\\
        CBMNet-L\cite{kim2023event} & \Checkmark & \XSolidBrush&  37.69/0.970 &37.57/0.983&36.04/0.976&36.99/0.974&35.32/0.962&33.63/0.949&31.84/0.931\\
        DSEVFI\cite{liu2024video} & \Checkmark & \XSolidBrush&   \underline{39.17}/0.977&36.95/0.975&35.77/0.968&-&-&33.04/0.914&31.46/0.893\\
        TLXNet+\cite{ma2024timelens} & \Checkmark & \XSolidBrush&   -&38.06/\underline{0.984}&\textbf{36.22}/\underline{0.977}&37.85/0.976&35.98/0.964&34.22/0.953&\underline{32.20}/0.935\\
        \addlinespace
        \midrule
        Ours-f3 & \Checkmark & \Checkmark&  \textbf{41.15}/\textbf{0.990}&37.54/\underline{0.984}&35.44/0.975&\underline{40.13}/\underline{0.990}&\textbf{37.23}/\textbf{0.980}&\underline{34.67}/\textbf{0.967}&32.05/\underline{0.946} \\
        Ours-f4 & \Checkmark & \Checkmark&  \underline{40.92}/\textbf{0.990}& \textbf{38.14}/\textbf{0.986}&\underline{36.20}/\textbf{0.978}&39.72/0.986&\underline{37.04}/0.975&\textbf{34.71}/\underline{0.963}&\textbf{32.54}/\textbf{0.949} \\
           \addlinespace
        \bottomrule
    \end{tabular} \label{Synthetic Event Datasets}
\begin{tablenotes}
\footnotesize
\item[1] The best and second-best results on each dataset are highlighted in bold and underlined, respectively. The performance of TimeLens, CBMNet-L and TLXNet+ are derived by the finetuned models based on our simulated data and implementation in \cite{ma2024timelens}.
\end{tablenotes}
\end{table*}

\subsection{Evaluation on Synthetic Event VFI Datasets}

Table \ref{Synthetic Event Datasets}  presents the performance of VFI methods on synthetic event VFI datasets, including Vimeo90k-Triplet, GoPro and SNU-FLIM. We compare our approach with SOTA VFI methods across three categories: frame-based, diffusion-based, and event-based VFI. Specifically, ``f3" denotes that the HAE employs three downsampling blocks, resulting in the smallest feature representation $\mathcal{F}^s_{k}(I)$ and $\mathcal{F}^s_{k}(E)$ with the spatial dimension of $\frac{H}{8} \times \frac{W}{8}$. Meanwhile, ``f4" utilizes four downsampling blocks, producing the smallest feature maps at $\frac{H}{16} \times \frac{W}{16}$. 

\begin{figure*}[!t]
\centering
\includegraphics[width=\textwidth]{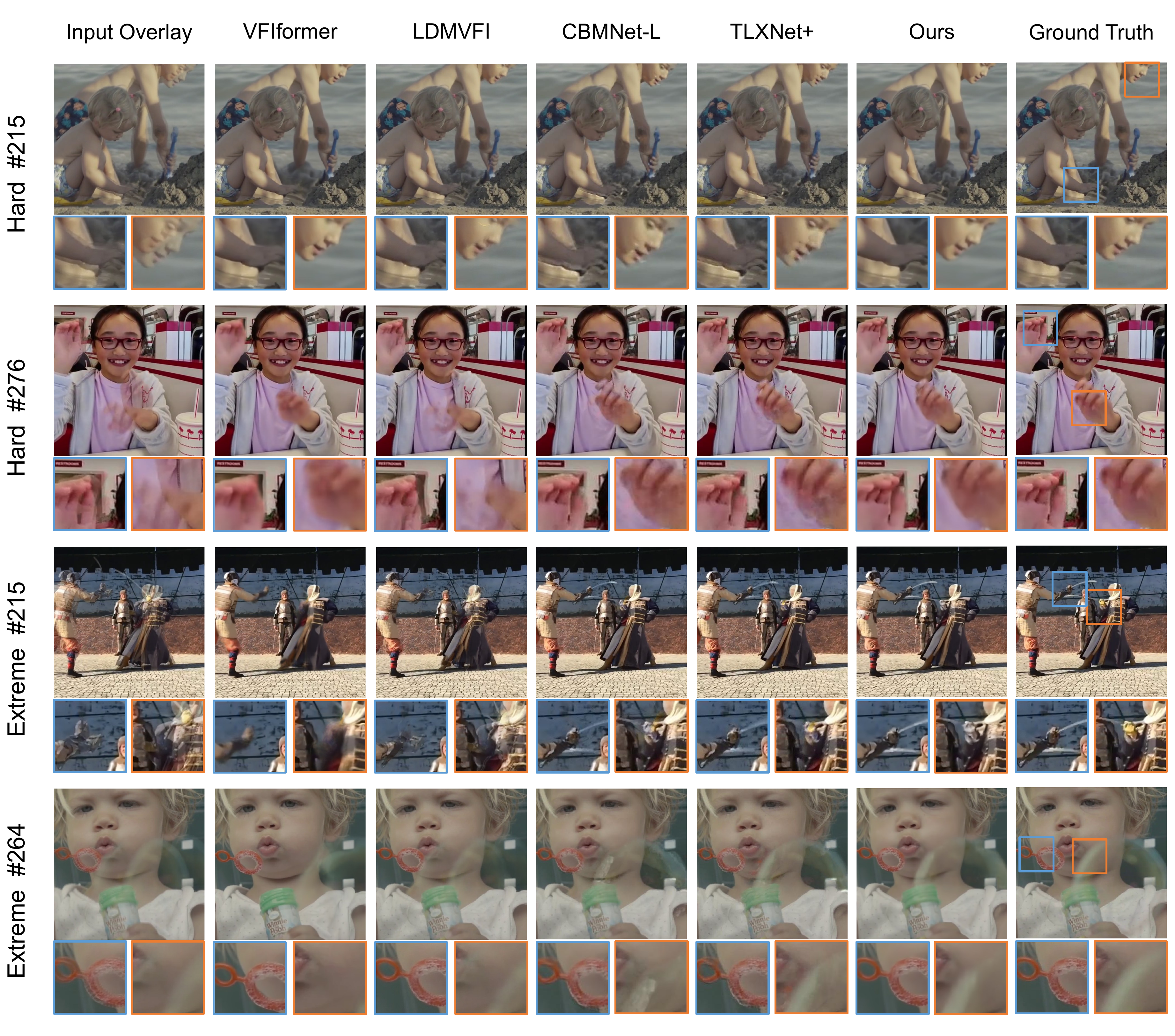}
\caption{Visual comparison of VFI methods on the SUN-FILM dataset with hard-level (the first two rows) and extreme-level (the last two rows) conditions. We select two sub-regions to zoom-in for better detail. The numbers on the vertical axis indicate the sample indices in the dataset.}
\label{visual_synthetic}
\end{figure*}

As demonstrated in Table \ref{Synthetic Event Datasets}, our method consistently outperforms existing approaches across these datasets in terms of PSNR and SSIM metrics. Specifically, on the Vimeo90k-Triplet dataset, our method achieves an improvement of 1.98dB in PSNR and 0.013 in SSIM compared to the best-performing event-based VFI method. Additionally, it surpasses the leading frame-based method by 4.51dB in PSNR and 0.008 in SSIM. Compared to the emerging diffusion-based VFI approach, our method achieves up to 5.72dB PSNR gain. On the GoPro dataset where we evaluate our method by skipping 7 and 15 frames, our method consistently demonstrates superior accuracy in the 7-skip setting and achieves comparable performance in the 15-skip setting across all comparison baselines.

The SNU-FLIM dataset, which poses greater challenges by incorporating varying levels of difficulty, provides a comprehensive evaluation benchmark for interpolation robustness. Our experiments indicate that frame-based methods perform well under easy and medium settings, whereas previous event-based methods demonstrate superior performance under hard and extreme conditions. This discrepancy arises because event-based methods are designed to recover large and complex motions via flow estimation and warping, often at the expense of high-fidelity image details when motion changes are relatively minor. Conversely, frame-based methods struggle with large motion interpolation due to the lack of explicit motion guidance. 

However, our method consistently exhibits strong performance across all difficulty levels. For instance, the SOTA event-based method TLXNet+ \cite{ma2024timelens} achieves comparable PSNR results under extreme settings but performs significantly worse under easier conditions. Moreover, our approach demonstrates a notable advantage in SSIM, as it directly reconstructs frames without relying on intermediate flow estimation, thereby mitigating propagation errors and structural distortions.

Furthermore, we observe that the Ours-f3 model performs exceptionally well in scenarios where motion changes between frames are moderate, such as the easy and medium settings on SNU-FLIM. In contrast, Ours-f4 demonstrates superior performance in extreme settings, where large and complex motions are present. This difference arises from the varying compression ratios of the spatial dimensions in HAE. A higher compression ratio enhances the model’s ability to capture large-scale motion dynamics, whereas a lower compression ratio preserves finer details, leading to more accurate reconstructions when motion changes are minimal.

\begin{table}[!htbp]
    \centering
    \caption{Performance (LPIPS$\downarrow$) comparison on perception quality with diffusion-based VFI methods.}
    \begin{tabular}{ccccc}
        \toprule
        \multirow{2}{*}{Method}  &  \multicolumn{2}{c}{SNU-FILM} &\multirow{2}{*}{Runtime (sec)}&\multirow{2}{*}{Params (M)}\\  \cmidrule(r){2-3}
       &hard&extreme&&\\
        \midrule

        VFIformer\cite{lu2022video} &0.061 &0.119 & 6.57&24.1\\
            \addlinespace
        CBMNet-L\cite{kim2023event}&0.060&0.079&9.33&22.2\\
            \addlinespace
        MCVD\cite{voleti2022mcvd}&0.250&0.320&36.23&27.3\\
            \addlinespace
        LDMVFI\cite{danier2024ldmvfi} & 0.060& 0.123&5.88&416.5\\
            \addlinespace
        BBD\cite{lyu2024frame} & \textbf{0.047}&0.104&1.74&109.9\\

        \midrule
        Ours-f3 & 0.053&\underline{0.085}&\textbf{0.38}&18.7\\
            \addlinespace
        Ours-f4 &\underline{0.052} &\textbf{0.076}&\underline{0.41}&41.7 \\
        \bottomrule
    \end{tabular} \label{LPIPS}
\begin{tablenotes}
\footnotesize
\item[1] The best and second-best results on each dataset are highlighted in bold and underlined, respectively.
\end{tablenotes}
\end{table}

Compared to frame-based and event-based methods, previous diffusion-based VFI approaches primarily focus on optimizing the perceptual quality of interpolated frames. However, they tend to underperform in terms of similarity metrics such as PSNR and SSIM. Our method is the first to leverage the diffusion model while achieving superior performance on these similarity metrics. To further assess the perceptual quality, Table \ref{LPIPS} presents the LPIPS scores of interpolated frames produced by three diffusion-based methods. Additionally, we include comparisons with a frame-based method (VFIformer) and an event-based method (CBMNet-L). The results demonstrate that our approach achieves high perceptual quality compared to other diffusion-based methods. Moreover, our method significantly improves the sampling efficiency. We measure the average running time needed to interpolate a frame at a resolution of $1280\times720$ on an NVIDIA RTX 4090 GPU, and the results indicate that our approach substantially outperforms all other diffusion-based methods. Specifically, our method achieves 4.24$\times$ faster inference compared with the SOTA diffusion-based methods. This efficiency gain is primarily due to our reduction in sampling iterations, i.e., from 200 steps in LDMVFI to just 5 steps in our method. Furthermore, HAE in our model has significantly fewer parameters compared to the autoencoder architectures used in other diffusion-based approaches, contributing to both improved performance and efficiency.

Fig. \ref{visual_synthetic} illustrates the VFI samples generated by various methods on the SNU-FILM dataset under hard and extreme difficulty levels. The frame-based method, VFIformer, struggles with visual interpolation due to the absence of sufficient motion information between frames, resulting in suboptimal performance. Diffusion-based approaches, such as LDMVFI, produce outputs resembling a simple overlay of the previous and next frames. These results indicate that solely optimizing the perceptual quality is insufficient for achieving satisfactory interpolation results. In contrast, event-based methods like CBMNNet-L and TLXNet+ demonstrate improved performance in handling large and complex motions. However, errors in flow estimation still lead to noticeable artifacts in the generated frames. Our method synthesizes the intermediate frame directly through a denoising process, eliminating the need for additional flow estimation. This approach effectively avoids potential flow-related errors and yields high-quality results with accurate reconstruction under large motion and finer detail preservation under moderate changes.

\begin{table*}[!htbp]
    \centering
    \caption{Quantitative performance comparison (PSNR $\uparrow$ (DB) / SSIM$\uparrow$) of results on real-world event VFI datasets.}\label{real}
    \begin{tabular}{ccccccccc}
        \toprule
        \multirow{2}{*}{Method} & \multirow{2}{*}{Events} & \multirow{2}{*}{DM}  & \multicolumn{2}{c}{HQ-EVFI}& \multicolumn{2}{c}{BS-ERGB} & \multicolumn{2}{c}{HQF} \\ \cmidrule(r){4-5} \cmidrule(r){6-7} \cmidrule(r){8-9}
        &&&7-skips&15-skips&1-skips&3-skips&1-skips&3-skips\\
        \midrule

        RIFE\cite{huang2022real} & \XSolidBrush & \XSolidBrush&25.59/0.920&21.34/0.863&25.39/0.808&22.67/0.763&31.70/0.889&27.93/0.796\\
        UNP-Net-L\cite{jin2023unified} & \XSolidBrush & \XSolidBrush&26.67/0.933&21.90/0.882&25.85/0.821&23.15/0.779&31.19/0.887&26.72/0.776\\
        VFIformer\cite{lu2022video} & \XSolidBrush & \XSolidBrush& 25.66/0.926 &21.30/0.873&25.70/0.821&22.86/0.778&31.02/0.886&26.50/0.772\\
        EMA-L\cite{zhang2023extracting} & \XSolidBrush & \XSolidBrush& 27.10/0.933&22.18/0.882&25.86/0.820&23.20/0.778&30.82/0.885&26.57/0.779\\
        \addlinespace
        \midrule
        LDMVFI\cite{danier2024ldmvfi} & \XSolidBrush & \Checkmark& 24.52/0.913&21.25/0.868 &24.44/0.790&22.07/0.749&29.89/0.853&25.67/0.738\\
        BBD\cite{lyu2024frame} & \XSolidBrush & \Checkmark& 25.26/0.924 &21.22/0.872&24.93/0.802&22.24/0.769&30.37/0.861&26.14/0.754\\
        \addlinespace
        \midrule
        TimeLens\cite{tulyakov2021time} & \Checkmark & \XSolidBrush& 34.37/\underline{0.978} &32.77/0.971&28.57/0.843&27.34/0.821&33.42/0.934&32.27/0.917\\
        A2OF\cite{wu2022video} & \Checkmark & \XSolidBrush& -&-&-&-&33.94/0.945&31.85/0.932\\
        DSEVFI\cite{liu2024video} & \Checkmark & \XSolidBrush&-&-&-&-&\underline{35.89}/\textbf{0.959}&\underline{34.27}/\underline{0.941}\\
        CBMNet-L\cite{kim2023event} & \Checkmark & \XSolidBrush& \underline{34.60}/\underline{0.978} &32.95/0.970&29.16/\underline{0.850}&28.42/0.843&34.77/0.953&33.08/0.940\\
        TLXNet+\cite{ma2024timelens}& \Checkmark & \XSolidBrush& 34.40/\underline{0.978} &\underline{33.13}/\underline{0.974}&\textbf{29.35}/\underline{0.850}&\textbf{28.79}/\underline{0.844}&-&-\\
        \addlinespace
        \midrule
        
        Ours & \Checkmark & \Checkmark&  \textbf{35.32}/\textbf{0.981}&\textbf{33.49}/\textbf{0.975}&\textbf{29.35}/\textbf{0.856}&\underline{28.49}/\textbf{0.849}&\textbf{36.31}/\textbf{0.959}&\textbf{34.44}/\textbf{0.943} \\
        \bottomrule
    \end{tabular}
\begin{tablenotes}
\footnotesize
\item[1] The best and second-best results on each dataset are highlighted in bold and underlined, respectively.
\end{tablenotes}
\end{table*}

\subsection{Evaluation on Real-world Event VFI Datasets}

To evaluate the effectiveness of our approach on real-world event VFI datasets, we further test our method on three event VFI datasets: HQ-EVFI, BS-ERGB, and HQF. These datasets are specifically designed to highlight the advantages of event-based VFI methods, as they involve large and complex motion variations as well as significant brightness fluctuations. In addition, the event data and frames in these datasets are collected with DVS and frame cameras in real-world scenarios.

Table \ref{real} presents a comparative analysis of our approach against other SOTA methods. Here we report only the performance of Our-f4 as it probably outperforms Our-f3 on datasets characterized by large and complex motion. The results demonstrate that our method consistently achieves superior performance on all datasets. Notably, while TLXNet+ achieves a higher PSNR than our approach on BS-ERGB, this can be partially attributed to the significant noise in BS-ERGB images \cite{ma2024timelens}, which reduces the reliability of PSNR as a metric. Despite a slightly lower PSNR, our method achieves the highest SSIM result on BS-ERGB, which is a more robust indicator of image quality, especially in noisy conditions. In contrast to BS-ERGB, the HQ-EVFI dataset features higher-quality images. On this dataset, our method surpasses all competing methods in terms of both PSNR and SSIM metrics, further confirming its effectiveness in real-world event-based VFI scenarios. Furthermore, on the HQF dataset, our method also achieves SOTA performance, providing additional evidence of its robustness and generalization on diverse real-world event VFI datasets.

\begin{figure*}[!t]
\centering
\includegraphics[width=\textwidth]{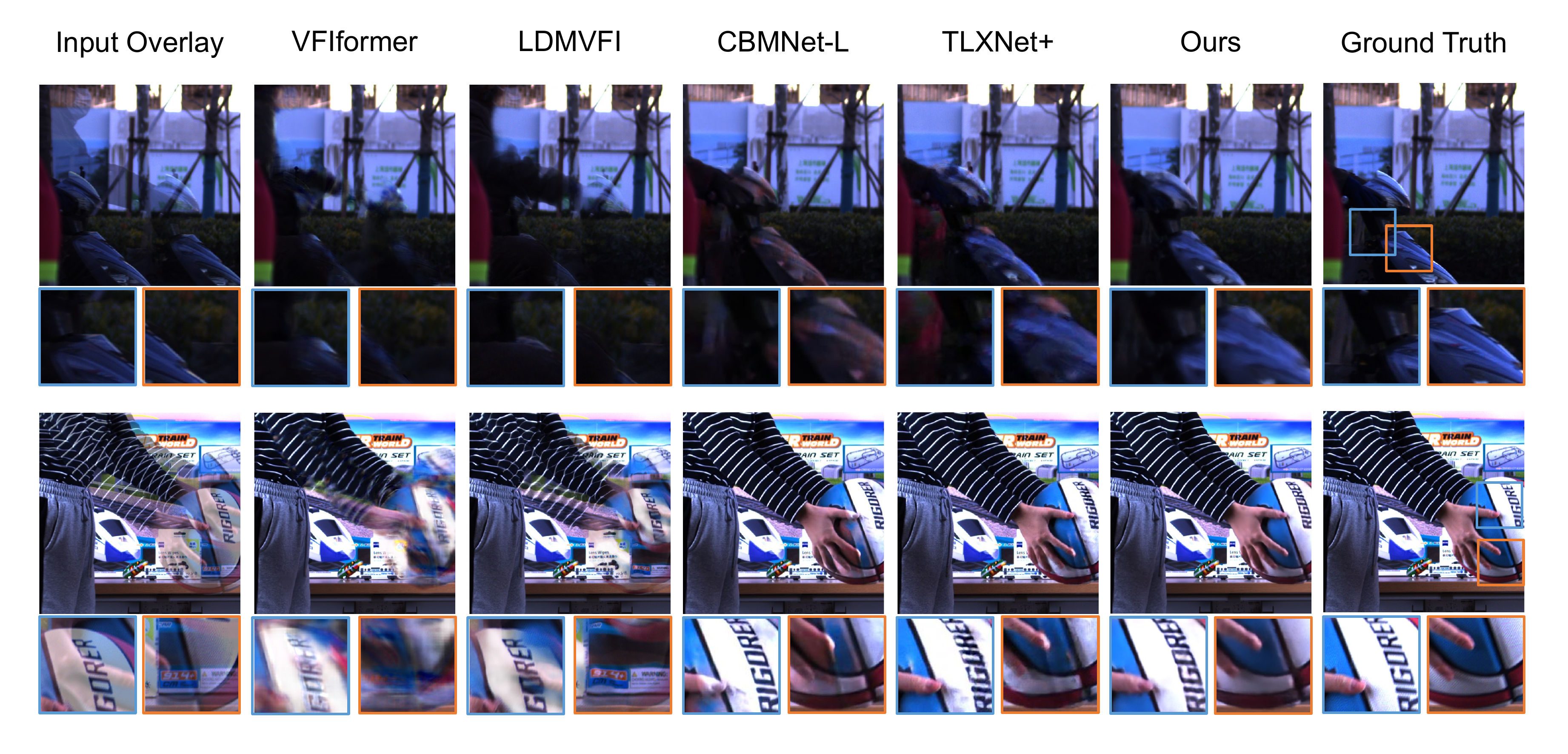}
\caption{Visual comparison of VFI methods on HQ-EVFI dataset. We select two sub-regions to zoom-in for better details.}
\label{visual-real}
\end{figure*}

Fig. \ref{visual-real} presents the VFI results produced by various methods on the HQ-EVFI dataset. Compared to synthetic datasets, traditional frame-based and diffusion-based methods exhibit noticeably poorer performance, primarily due to the presence of larger and more non-linear motions in real-world event VFI datasets. Although event-based methods show  improved results, they still suffer from noticeable artifacts and color distortions in challenging regions. In contrast, our method demonstrates superior visual quality, effectively handling complex motion and preserving image fidelity better than competing approaches.

\subsection{Model Analysis}

\subsubsection{The spatial size of the ground-truth embedding} In this subsection, we analyze the impact of the spatial size of the ground-truth embedding on the model performance. From previous experiments, we observed that Our-f3 and Our-f4 models, which use different ground-truth embedding sizes, exhibit varying performance levels. To further investigate this, we conduct additional experiments on several model variants, including Our-f2 and Our-f5, to comprehensively examine how the spatial size of the ground-truth embedding affects performance during training Stages 1 and 2.

\begin{table}[!htbp]
    \centering
    \caption{Performance (PSNR $\uparrow$ (DB) / SSIM$\uparrow$) analysis on the impact of the spatial size of the ground-truth embedding.} \label{size}
    \begin{tabular}{ccccc}
        \toprule
        \multirow{2}{*}{Model}  & \multirow{2}{*}{Spatial size} & \multicolumn{2}{c}{Vimeo90K-Triplet}&\multirow{2}{*}{Params (M)}\\ \cmidrule(r){3-4}
        &&Stage 1 & Stage 2 &\\
        \midrule

        Our-f2&$\frac{H}{4}\times\frac{W}{4}$&\textbf{42.91}/\textbf{0.992}&40.49/\underline{0.989}&17.9 \\
        \addlinespace
        Our-f3&$\frac{H}{8}\times\frac{W}{8}$&\underline{41.62}/\underline{0.991}&\textbf{41.15}/\textbf{0.990}&18.7\\
        \addlinespace
        Our-f4&$\frac{H}{16}\times\frac{W}{16}$&41.00/0.990&\underline{40.92}/\textbf{0.990}&41.7\\
        \addlinespace
        Our-f5&$\frac{H}{32}\times\frac{W}{32}$&40.62/0.989&40.61/\underline{0.989}&45.3\\
    \addlinespace
        \bottomrule
    \end{tabular}
\begin{tablenotes}
\footnotesize
\item[1] The best and second-best results  are highlighted in bold and underlined, respectively.
\end{tablenotes}
\end{table}

Each model variant, denoted as Our-fn, contains $n$ downsampling blocks in HAE, resulting in a ground-truth embedding spatial size of $\frac{H}{2^n}\times\frac{W}{2^n}$ relative to the original image size $H\times W$. This means that models with smaller ground-truth embeddings capture less spatial information from the ground-truth image in Stage 1 during training, leading to weaker performance at this stage. Table \ref{size} presents the PSNR and SSIM results of each model variant on the Vimeo90K-Triplet dataset. Notably, models with a larger spatial size of the ground-truth embedding achieve better performance with fewer parameters after Stage 1 during training. However, increasing the spatial size of the ground-truth embedding can negatively affect performance in Stage 2. A larger embedding makes the reconstruction process more challenging during the diffusion process, leading to greater degradation in performance. As shown in Table \ref{size}, Our-f4 and Our-f5 models, which have smaller ground-truth embeddings, experience minimal performance loss during Stage 2, as the primary bottleneck occurs in Stage 1. Conversely, Our-f2 and Our-f3 models suffer from performance degradation in Stage 2 due to the difficulty in reconstructing high-quality image embeddings during the diffusion process. Therefore, achieving optimal final performance requires selecting an appropriate spatial size for the ground-truth embedding, striking a balance between the trade-offs in Stages 1 and 2.

\begin{table}[!htbp]
    \centering
    \caption{Performance comparison of different fusion methods in the encoder of HAE.} 
    \begin{tabular}{ccccc}
        \toprule
        \multirow{2}{*}{Fusion methods}   & \multicolumn{3}{c}{Vimeo90K-Triplet}&\multirow{2}{*}{Params (M)}\\ \cmidrule(r){2-4}
        &$\mathcal{L}_1$&PSNR & SSIM &\\
        \midrule
        Concat&0.01204&40.82&0.989&5.2 \\    
        \addlinespace
        Concat-L&0.01123&41.46&0.990&6.1 \\
        \addlinespace             
        SCA&0.01142&41.28&0.990&5.9\\
        \addlinespace
        TCA&0.01187&40.94&0.989&5.4\\
        \addlinespace
        STCA&0.01098&41.62&0.991&6.0\\
   
    \addlinespace
        \bottomrule
    \end{tabular} \label{fusion}
\end{table}

\subsubsection{Analysis of the fusion method} We conduct an ablation study on the fusion methods used for integrating event data and image frames within the encoder of HAE during the training of Stage 1. We evaluate five different fusion strategies. (1) Concat: This method directly concatenates the event and frame inputs at the beginning of the HAE encoder without any additional attention-based fusion operations.(2) Concat-L: This method is similar to Concat, but with an increased number of parameters to match the model size of STCA, enabling a fairer comparison. (3) SCA (Spatial Cross-Attention): This approach employs only spatial cross-attention for modality fusion. (4) TCA (Temporal Cross-Attention): This method applies only temporal cross-attention for modality fusion. (5) STCA (Spatial-Temporal Cross-Attention, our proposed method): This method factorizes the fusion process into two distinct operations, SCA and TCA, to enhance the efficiency and effectiveness of modal integration.

Table \ref{fusion} presents the validation $\mathcal{L}_1$ loss, PSNR, and SSIM for the different fusion methods used in the HAE encoder of the Our-f3 model after Stage 1 during training. The results show that incorporating either the light-weighted SCA or TCA mechanism leads to performance improvement compared to simple concatenation. Notably, the proposed STCA method achieves the best performance, even surpassing the Concat-L variant, which has a larger model size. This ablation study clearly demonstrates the effectiveness of the proposed STCA fusion strategy.

\begin{table*}[!htbp]
    \centering
    \caption{Performance (average PSNR $\uparrow$ (DB) and SSIM $\uparrow$) of different trainings schemes of the diffusion model in Stage 2.}\label{diffusion}
    \begin{tabular}{l|cc|cccc|ccc}
        \toprule
        \multirow{3}{*}{Model} & \multirow{3}{*}{GT} & \multirow{3}{*}{DM}  & \multicolumn{4}{c|}{Training schemes}& \multirow{3}{*}{PSNR} & \multirow{3}{*}{SSIM} & \multirow{3}{*}{Params (M)}\\ \cmidrule(r){4-7} 
        &&&\makecell{Traditional \\ DM optimization}&\makecell{Cond. encoder \\ optimization}&\makecell{HAE encoder \\ optimization}&\makecell{HAE decoder \\optimization}&&&\\
        \midrule

        EventDiff-V0 & \Checkmark & \XSolidBrush & \XSolidBrush & \XSolidBrush & \XSolidBrush& \XSolidBrush&41.60&0.991&6.0\\
        \addlinespace
        \midrule
        EventDiff-V1 & \XSolidBrush & \XSolidBrush & \XSolidBrush & \XSolidBrush & \XSolidBrush& \XSolidBrush&34.93&0.976&6.0\\
                \addlinespace
        EventDiff-V2 & \XSolidBrush & \Checkmark & \Checkmark & \XSolidBrush & \XSolidBrush& \XSolidBrush&37.09&0.984&16.2\\
                \addlinespace
        EventDiff-V3 & \XSolidBrush & \Checkmark & \XSolidBrush& \XSolidBrush & \XSolidBrush& \XSolidBrush&38.36&0.986&16.2\\
                \addlinespace
        EventDiff-V4 & \XSolidBrush & \Checkmark & \XSolidBrush & \Checkmark & \XSolidBrush& \XSolidBrush&40.90&0.989&18.7\\
                \addlinespace
        EventDiff-V5(Ours) & \XSolidBrush & \Checkmark & \XSolidBrush & \Checkmark & \XSolidBrush& \Checkmark&\textbf{41.15}&\textbf{0.990}&18.7\\                  \addlinespace
        EventDiff-V6 & \XSolidBrush & \Checkmark & \XSolidBrush & \Checkmark & \Checkmark& \Checkmark&\underline{41.05}&\underline{0.990}&18.7\\   
                \addlinespace
        \bottomrule
    \end{tabular}
\begin{tablenotes}
\footnotesize
\item[1] The best and second-best results  are highlighted in bold and underlined, respectively.
\end{tablenotes}
\end{table*}

\subsubsection{Training scheme of the diffusion model in Stage 2} In this subsection, we conduct experiments to analyze the training schemes of Stage 2 in the diffusion process. We construct seven model variants, ranging from EventDiff-V0 to EventDiff-V7, based on the Ours-f3 model for comparative analysis. Among them, EventDiff-V0 is the pretrained model from Stage 1, which receives ground-truth guidance and serves as the foundation for subsequent model variants of the Our-f3 model developed in Stage 2. The details of the training scheme of EventDiff-V1 to EventDiff-V7 are outlined below:
\begin{itemize}
    \item EventDiff-V1 replaces the ground-truth embedding with the mean of the condition embeddings generated by an untrainable condition encoder, which takes boundary I-E pairs as input. No further optimization operations are applied.
    \item EventDiff-V2: employs a traditional diffusion model to approximate the ground-truth embedding using condition embeddings. This model estimates noise at each time step and generates the final denoised embedding $\hat{z}$ after thousands of iterations. The diffusion model follows the standard DDPM training process with 1000 time steps.
    \item EventDiff-V3, V4, V5, and V6 adopt the proposed method that directly estimates the final denoised embedding $\hat{z}$. The condition encoder in EventDiff-V3 remains untrainable, while it is optimized in other variants.
    \item EventDiff-V5 represents the training scheme we adopt, which jointly optimizes the condition encoder and the HAE decoder along with the diffusion model. Different from EventDiff-V5, EventDiff-V4 keeps the HAE decoder untrainable while the HAE encoder of EventDiff-V6 is trainable.
\end{itemize}

Table \ref{diffusion} presents the performance comparison under different training schemes in Stage 2. Compared to EventDiff-V1 and V2, EventDiff-V3 demonstrates superior performance, indicating that directly minimizing the error between the final estimated denoised embedding $\hat{z}$ and the ground-truth embedding $z_{gt}$ is significantly more effective than traditional diffusion model optimization. Moreover, EventDiff-V4 outperforms EventDiff-V3, highlighting that optimizing the condition encoder further enhances the model performance. The results of EventDiff-V5 and V6 provide deeper insights on the impact of joint optimization of the HAE modules. Specifically, jointly optimizing the HAE decoder (as in V5) improves performance, as it helps the model adapt to residual noise not eliminated by the diffusion model. In contrast, optimizing the HAE encoder (as in V6) has a slightly negative impact, likely because it disrupts the hybrid feature pyramid representation pretrained in Stage 1, thereby degrading the model performance.

\begin{figure}[!htbp]
\centering
\includegraphics[width=0.5\textwidth]{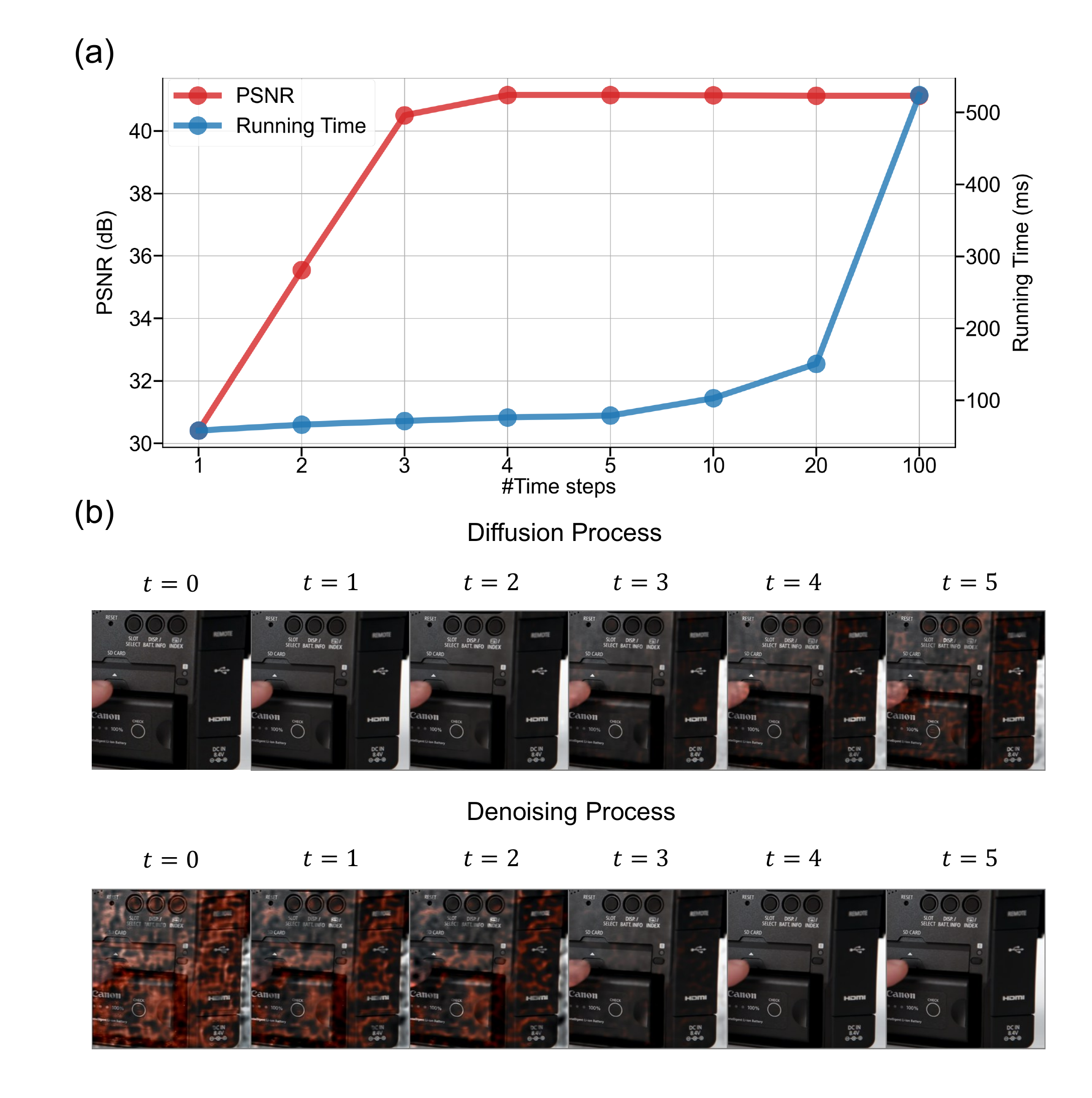}
\caption{Analysis of the iterative diffusion and denoising processes of the EventDiff: (a) The changes in VFI performance and running time as the number of iteration steps increases; (b) Visualization of the diffusion and denoising processes at different steps.}
\label{denoising}
\end{figure}

\subsubsection{Impact of the number of iteration steps} In this subsection, we analyze the impact of the number of iteration steps in the diffusion process on the model performance. As shown in Fig. \ref{denoising}(a), both the performance and running time of the diffusion model change as the number of iteration steps increases. The PSNR metric is evaluated on the Vimeo-Triplet dataset, and the running time refers to the average time required to interpolate a frame of size $448 \times 256$ using an NVIDIA RTX 4090 GPU. The results indicate that the VFI model achieves satisfactory performance after just four steps, reaching a PSNR result of 41.148dB, very close to the best result of 41.15dB obtained after five steps. The visualization in Fig. \ref{denoising}(b) further supports this observation, showing that most of the noise in the frame has been removed after four steps in the denoising process. These findings suggest that EventDiff can achieve competitive performance with a minimal number of diffusion iteration steps. Moreover, the interpolation of a single frame takes only 78.78ms after five steps, compared to the 524.25ms after 100 steps. Therefore, by significantly reducing the number of iteration steps, EventDiff can effectively lower the computational cost of frame interpolation.

\subsection{Extensibility Verification in Event Motion Deblurring Task}

In this subsection, we apply the method to the event motion deblurring task to verify the extensibility of EventDiff in other event-enhanced visual generation tasks. For demonstration, we use the widely adopted GoPro dataset \cite{nah2017deep}, which contains 3,214 pairs of high-resolution (1280$\times$720) blurry and sharp images. The blurry images are generated by averaging a sequence of high-speed sharp frames. Following the setup in EFNet \cite{sun2022event}, we use the ESIM simulator and the Symmetric Cumulative Event Representation (SCER) to generate corresponding event data for each blurry image.

To adapt EventDiff to the deblurring task, we make a few task-specific adjustments. Unlike the VFI setting where the encoder receives two boundary image-event (I-E) pairs, here we only provide a single I-E pair consisting of the blurry image and its associated event data. The ground-truth I-E pair comprises the sharp image and a padding event stream. Additionally, we replace the kernel-based synthesis module in the decoder, originally designed for VFI, with a simplified U-Net to better suit the deblurring objective. We adopt the f3 variant of EventDiff for this task.

As shown in Table \ref{deblurring}, our method outperforms the frame-based and diffusion-based baselines. Notably, although our model is not specifically designed for motion deblurring, it delivers performance comparable to the event-based deblurring methods specialized for such task. These results strongly validate the generality and flexibility of EventDiff, highlighting its potential as a unified framework for a broad range of event-enhanced visual generation tasks.

\begin{table}[!htbp]
    \centering
    \caption{Performance (PSNR $\uparrow$  / SSIM$\uparrow$) comparison of motion deblurring methods on GoPro.} \label{deblurring}
    \begin{tabular}{ccccc}
        \toprule
        \multirow{2}{*}{Method} & \multirow{2}{*}{Events} & \multirow{2}{*}{DM} & \multicolumn{2}{c}{GoPro} \\ \cmidrule(lr){4-5}
        & & & PSNR & SSIM \\ \midrule
SRN\cite{tao2018scale} & \XSolidBrush           & \XSolidBrush           & 30.26          & 0.934          \\
\addlinespace
HINet\cite{chen2021hinet} & \XSolidBrush           & \XSolidBrush           & 32.71          & 0.959          \\
\addlinespace
MSDI-Net\cite{li2022learning} & \XSolidBrush           & \XSolidBrush           & 33.28          & 0.964          \\
\addlinespace
NAFNet\cite{chen2022simple}& \XSolidBrush           & \XSolidBrush           & \underline{33.69}          & \underline{0.967}          \\
\addlinespace
UFPNet\cite{fang2023self} & \XSolidBrush           & \XSolidBrush           & \textbf{34.06}          & \textbf{0.968} \\ \midrule

D$^2$NET\cite{shang2021bringing} & \Checkmark          & \XSolidBrush           & 31.76          & 0.943          \\
\addlinespace
ERDNet\cite{haoyu2020learning}  & \Checkmark           & \XSolidBrush           & 32.99          & 0.935          \\
\addlinespace
EFNet\cite{sun2022event}  & \Checkmark           & \XSolidBrush           & 35.46          & 0.972          \\
\addlinespace

MAENet\cite{sun2024motion} & \Checkmark           & \XSolidBrush           & \underline{36.07}          & \underline{0.976}          \\
\addlinespace
STCNet\cite{yang2024motion} & \Checkmark           & \XSolidBrush           & \textbf{36.45}          & \textbf{0.978}          \\
\addlinespace
\midrule

DiffIR\cite{xia2023diffir} & \XSolidBrush           & \Checkmark           & 33.20          & 0.963          \\
\addlinespace
DPMs\cite{ren2023multiscale} & \XSolidBrush           & \Checkmark           & 33.20          & 0.963          \\
\addlinespace
HI-Diff\cite{chen2024hierarchical} & \XSolidBrush           & \Checkmark           & \underline{33.33}          & \underline{0.964}          \\
\addlinespace
Swintormer\cite{chen2024efficient}& \XSolidBrush           & \Checkmark           & \textbf{33.38}          & \textbf{0.965}          \\
\addlinespace
\midrule

Ours& \Checkmark           & \Checkmark           & 34.86 & 0.968          \\            \bottomrule
    \end{tabular}
\begin{tablenotes}
\footnotesize
\item[1] The best and second-best results in each kind of methods are highlighted in bold and underlined, respectively.
\end{tablenotes}
\end{table}

\section{Conclusion}

In this work, we propose EventDiff, a unified and efficient diffusion model framework for event-based VFI. EventDiff overcomes the limitations of traditional frame-based, event-based, and diffusion-based methods by leveraging a novel Event-Frame Hybrid AutoEncoder (HAE) with a Spatial-Temporal Cross Attention (STCA) module for effective multimodal fusion. Unlike prior event-based approaches that rely on complex warping/synthesis modules or handcrafted intermediate representations, our method performs global optimization directly in the latent space through a denoising diffusion process. 

By adopting a two-stage training strategy, pretraining HAE with ground-truth supervision and subsequently conducting joint optimization with the diffusion model, EventDiff achieves superior interpolation accuracy and significantly improved efficiency by dramatically reducing the number of required inference steps. Our experiments on both synthetic and real-world event VFI datasets demonstrate that EventDiff consistently outperforms SOTA frame-based, event-based, and diffusion-based VFI methods, exhibiting strong robustness across diverse and challenging scenarios. Specifically, EventDiff surpasses leading event-based VFI methods by up to 1.98dB in PSNR on the Vimeo90K-Triplet dataset and delivers comprehensive improvements across all difficulty levels on the SNU-FILM benchmark. Compared to emerging diffusion-based VFI approaches, our method achieves up to 5.72dB PSNR gain on Vimeo90K-Triplet and offers 4.24$\times$ faster inference speed. Additionally, we validate the extensibility of EventDiff by applying to the event-based motion deblurring task, further confirming its potential as a general-purpose framework for event-enhanced visual generation. We believe EventDiff opens up new possibilities for applying diffusion models to diverse  event-based vision tasks.


\begin{IEEEbiography}[{\includegraphics[width=1in,height=1.25in,clip,keepaspectratio]{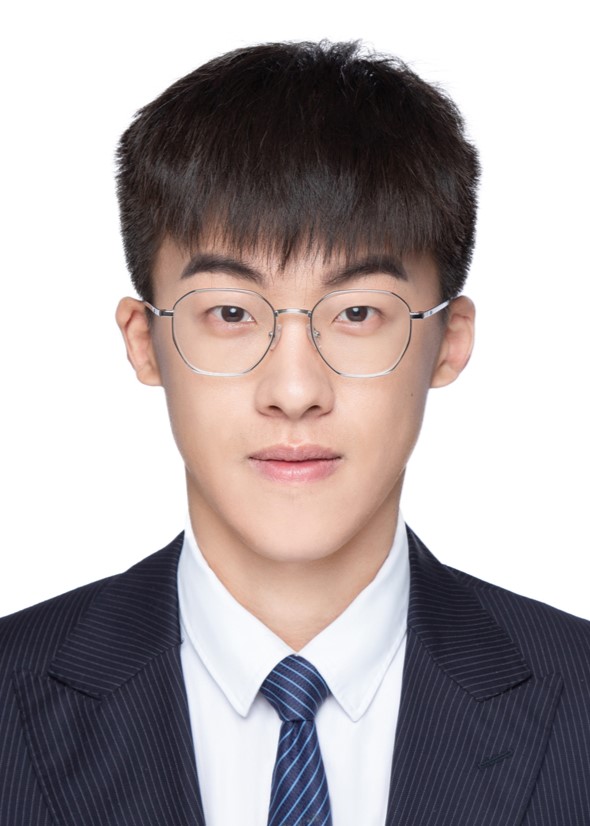}}]{Hanle Zheng} received the B.E. degree in Measurement and Control Technology and Instruments from the Department of Precision Instrument, Tsinghua University, Beijing, China, in 2022. He is currently pursuing the Ph.D. degree with the Department of Precision Instrument, Tsinghua University, Beijing, China. His research interests include brain-inspired learning, neuromorphic computing, spiking neual networks and dynamic system. \end{IEEEbiography}

\begin{IEEEbiography}[{\includegraphics[width=1in,height=1.25in,clip,keepaspectratio]{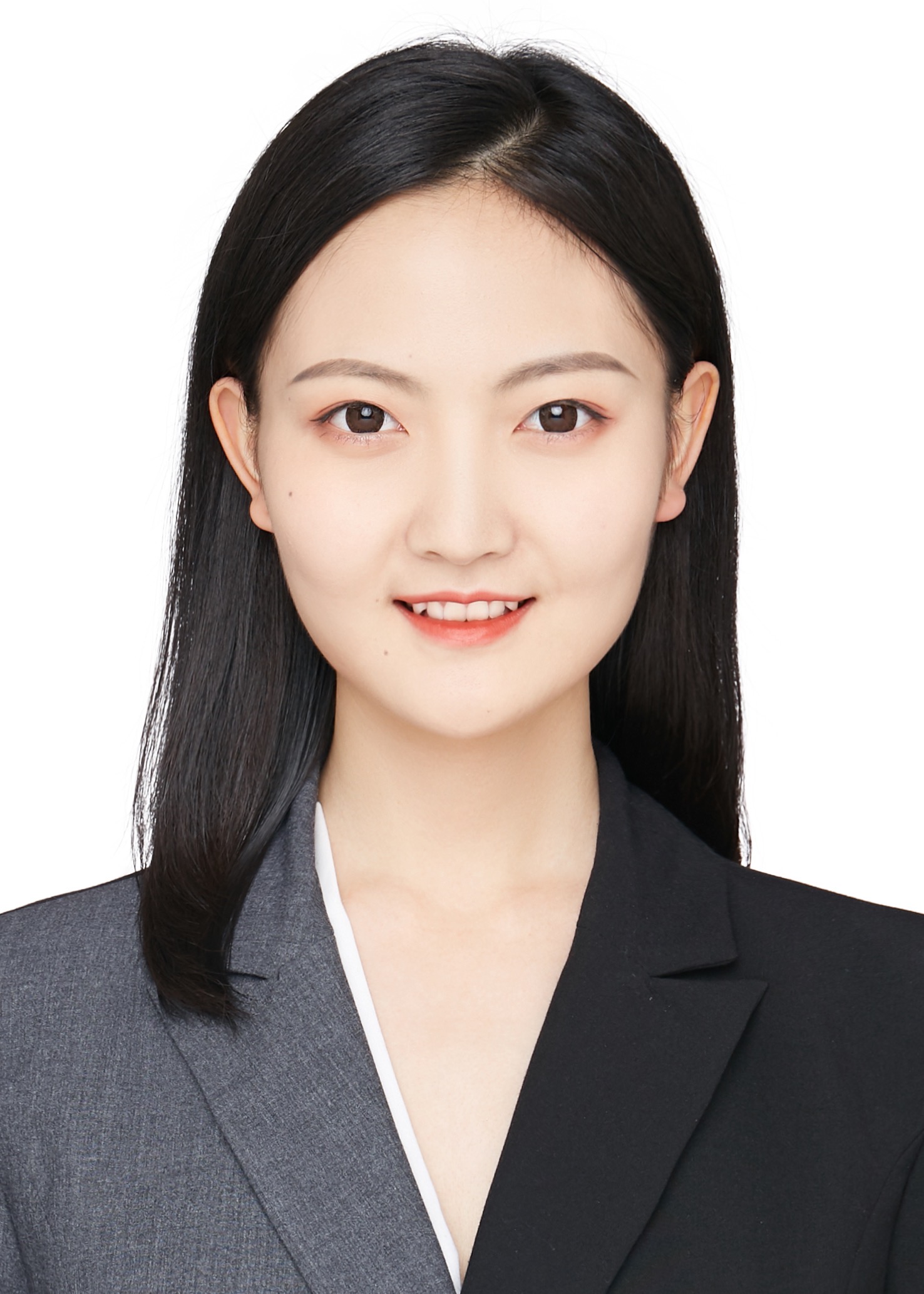}}]{Xujie Han} is currently pursuing the Ph.D. degree with the College of Computer Science and Technology, Taiyuan University of Technology, Jinzhong, China. Her research interests include brain-inspired computing, computer vision, deep learning, and their applications. \end{IEEEbiography}

\begin{IEEEbiography}[{\includegraphics[width=1in,height=1.25in,clip,keepaspectratio]{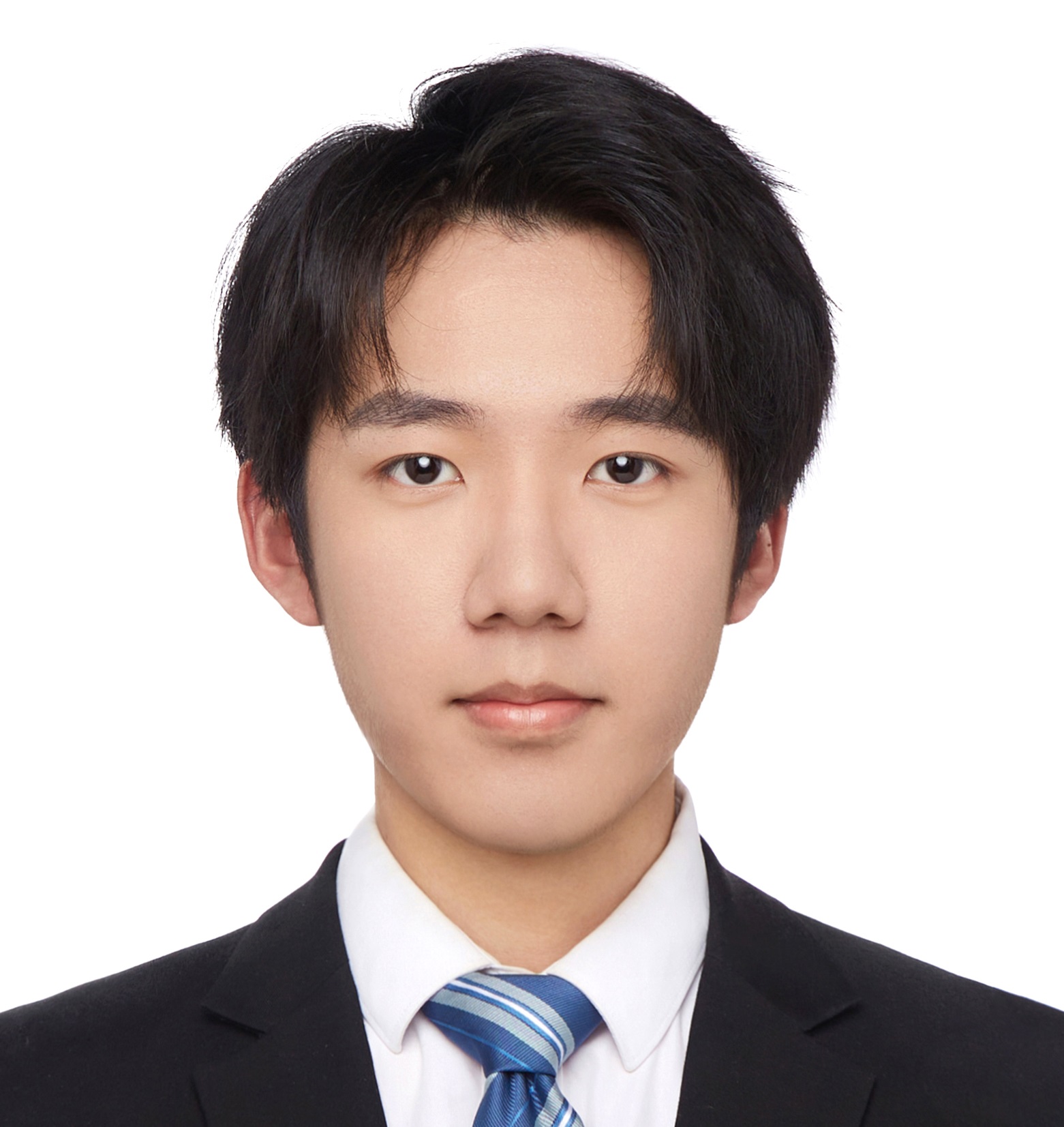}}]{Zegang Peng} received the B.E. degree in Communication Engineering from the School of Telecommunications Engineering, Xidian University, Shanxi, China, in 2024. He is currently pursuing the M.S. degree with the Department of Precision Instrument, Tsinghua University, Beijing, China. His research interests include brain-inspired computing, machine learning and neuromorphic chip. \end{IEEEbiography}
\begin{IEEEbiography}[{\includegraphics[width=1in,height=1.25in,clip,keepaspectratio]{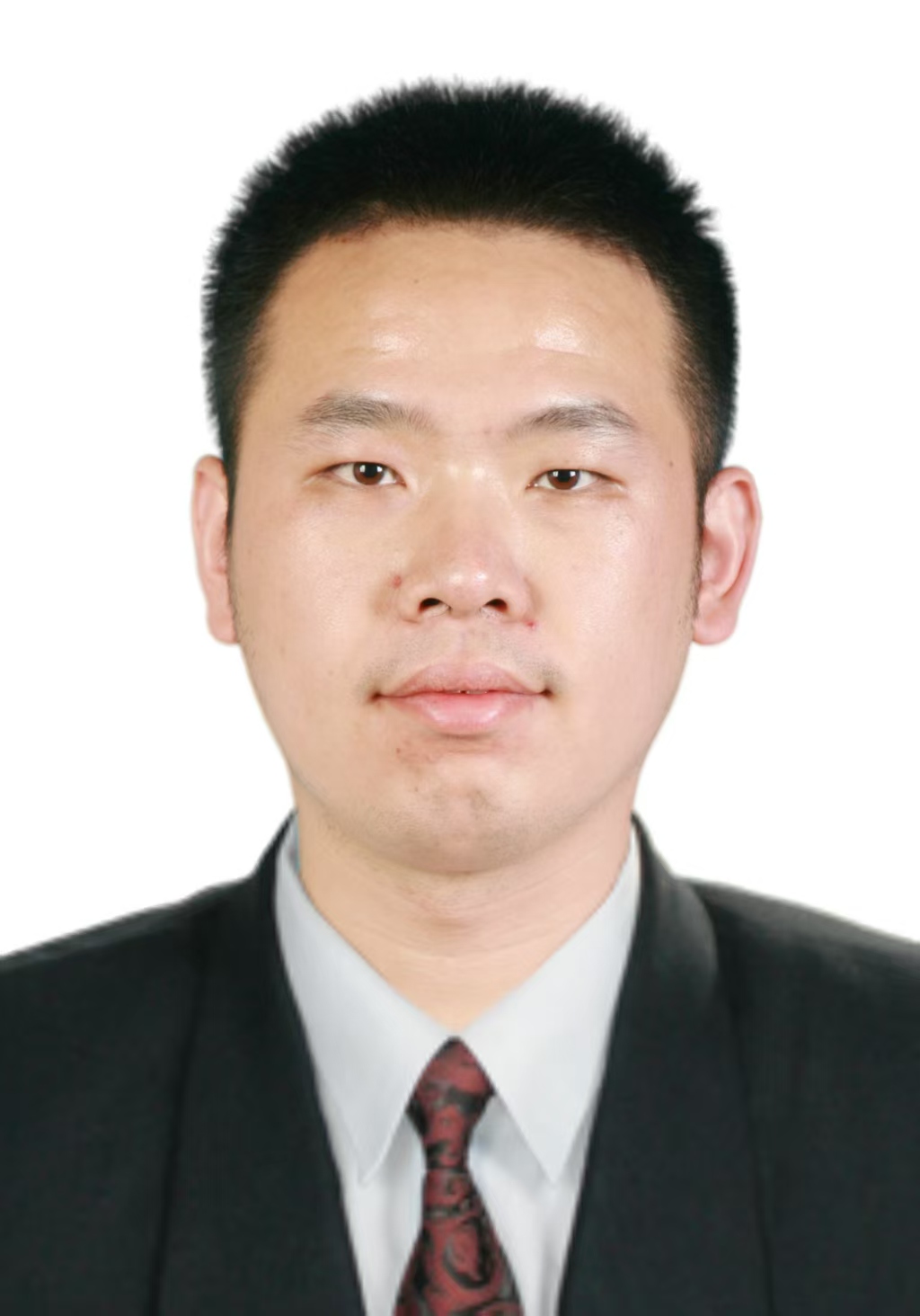}}]{Shangbin Zhang} received the B.E. and Ph.D. degrees from University of Science and Technology of China in 2012 and 2017, respectively. He was a Postdoctoral Fellow at Beihang University from 2017 to 2019. He is currently a Senior Engineer at the Information Science Academy of China Electronics Technology Group Corporation, Beijing, China. His primary research areas include swarm intelligence and cooperative decision-making and control. \end{IEEEbiography}
\begin{IEEEbiography}[{\includegraphics[width=1in,height=1.25in,clip,keepaspectratio]{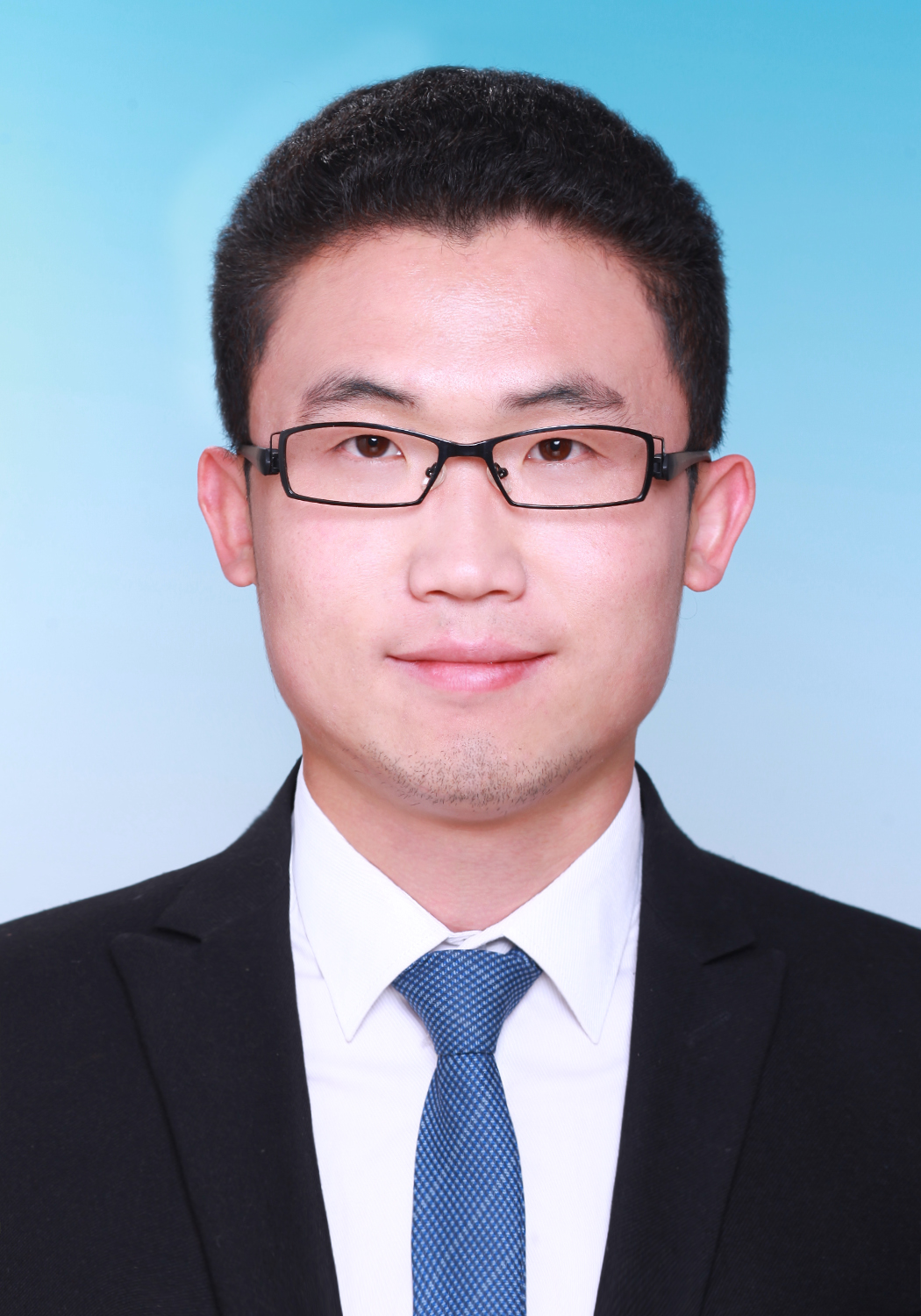}}]{Guangxun Du} received the B.E. and Ph.D. degrees from Beihang University, Beijing, China in 2009 and 2016, respectively. He was a Postdoctoral Fellow at Beihang University from 2016 to 2018. He is currently a Senior Engineer at the Information Science Academy of China Electronics Technology Group Corporation, Beijing, China. His primary research areas include autonomous intelligence, swarm intelligence, cooperative decision-making and control, and collaborative cluster intelligence. \end{IEEEbiography}

\begin{IEEEbiography}[{\includegraphics[width=1in,height=1.25in,clip,keepaspectratio]{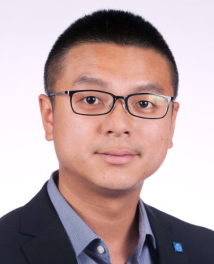}}]{Zhuo Zou} (Senior Member, IEEE) received the Ph.D. degree in electronic and computer systems from the KTH Royal Institute of Technology (KTH), Sweden, in 2012. Currently, he is with Fudan University, Shanghai, China, as a Full Professor, where he is conducting research on intelligent chips and systems for AIoT. Prior to joining Fudan, he was the Assistant Director and a Project Leader with the VINN iPack Excellence Center, KTH. He was an Adjunct Professor and a Docent with the University of Turku, Finland. His current research interests include low-power circuits, energy-efficient SoC, neuromorphic computing and their applications in AIoT, and autonomous systems. He is the Vice Chair of IFIP WG-8.12. \end{IEEEbiography}

\begin{IEEEbiography}[{\includegraphics[width=1in,height=1.25in,clip,keepaspectratio]{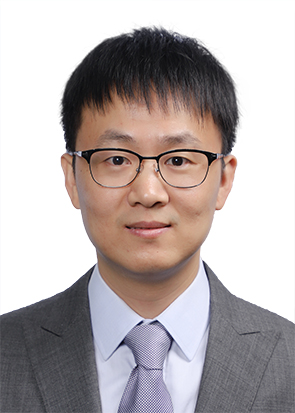}}]{Xilin Wang} received the B.E. degree in Materials Science and Technology from Department of Materials Science and Technology, Beijing University of Chemical Technology and the Ph.D. degree in Materials Science and Technology from Department of Materials Science and Technology, Tsinghua University. He is currently an Associate Professor with the Shenzhen Graduate School of Tsinghua University, Shenzhen, China. His primary research areas include high-voltgae insulation and dielectric materials, and the interdisciplinary fields between materials science , electrical engineering and intelligent technologies. \end{IEEEbiography}

\begin{IEEEbiography}[{\includegraphics[width=1in,height=1.25in,clip,keepaspectratio]{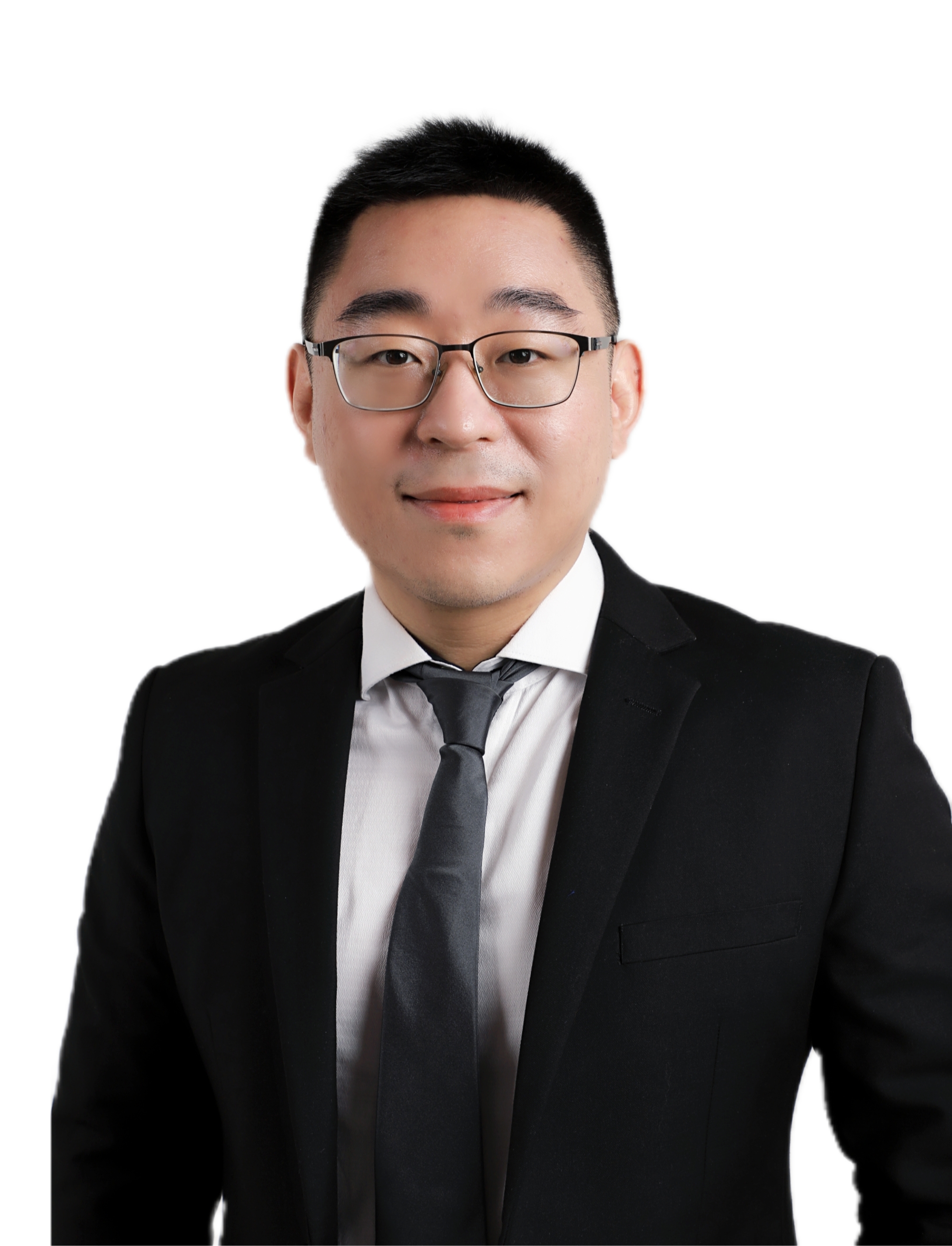}}]{Jibin Wu} (Member, IEEE) received the B.E. and Ph.D. degree in Electrical Engineering from National University of Singapore, Singapore in 2016 and 2020, respectively. He is currently an Assistant Professor in the Department of Data Science and Artificial Intelligence, The Hong Kong Polytechnic University. His research interests broadly include brain-inspired artificial intelligence, neuromorphic computing, computational audition, speech processing, and machine learning. He has published over 40 papers in prestigious conferences and journals in artificial intelligence and speech processing, including NeurIPS, ICLR, AAAI, TPAMI, TNNLS, TASLP, and IEEE JSTSP. He is currently serving as the Associate Editors for IEEE Transactions on Neural Networks and Learning Systems and IEEE Transactions on Cognitive and Developmental Systems.\end{IEEEbiography}

\begin{IEEEbiography}[{\includegraphics[width=1in,height=1.25in,clip,keepaspectratio]{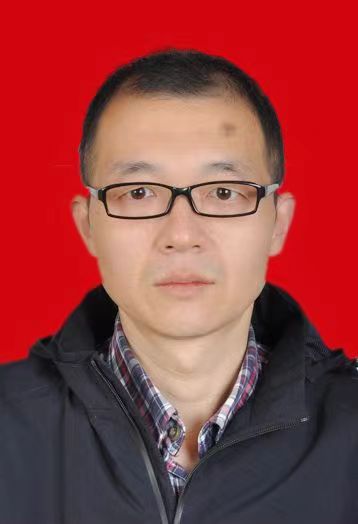}}]{Hao Guo} received the Ph.D. degree from TaiYuan University of Technology, Taiyuan, China in 2013. He was a visiting scholar at Nanyang Technological University, Singapore in 2018. He is currently a Full Professor at TaiYuan University of Technology. He has focused on studying brain-inspired models, brain informatics, and their applications for more than 15 years. \end{IEEEbiography}

\begin{IEEEbiography}[{\includegraphics[width=1in,height=1.25in,clip,keepaspectratio]{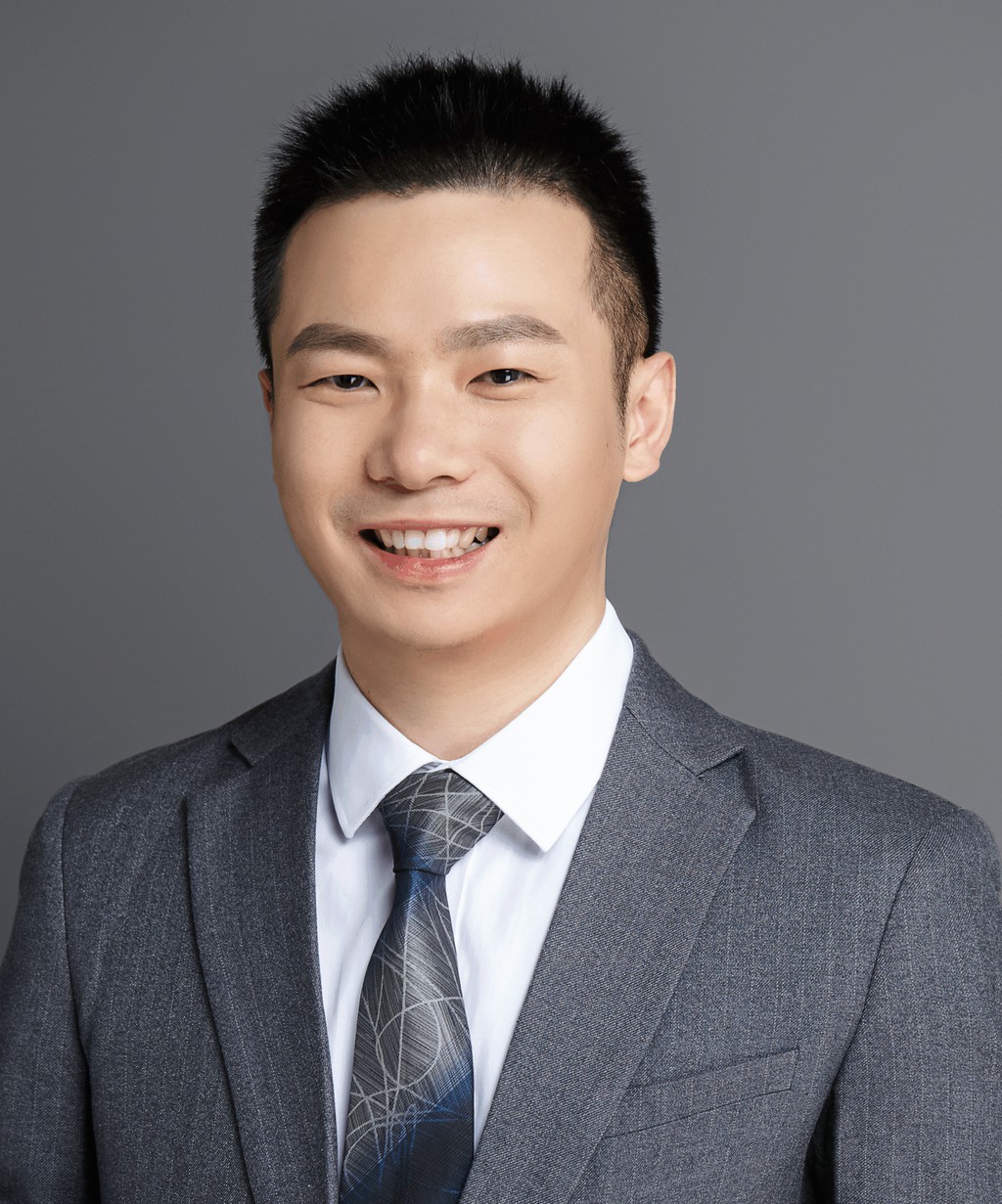}}]{Lei Deng} (Senior Member, IEEE) received the B.E. degree from University of Science and Technology of China in 2012 and the Ph.D. degree from Tsinghua University in 2017. He was a Postdoctoral Fellow at University of California, Santa Barbara from 2017 to 2021. He is currently an Associate Professor at Tsinghua University. He has focused on studying brain-inspired models, chips, and applications for more than 12 years. He has published over 100 papers in prestigious journals/conferences such as Nature, Nature Communications, Proc. IEEE, IEEE TPAMI, and IEEE JSSC, including four Cover Papers, five ESI Highly-Cited papers, and one Best Paper. He was a recipient of the Young Scholar of Chinese Institute for Brain Research, Beijing, Outstanding Youth Award of Chinese Association for AI, MIT TR 35 China, and Intel China Academic Achievement Award (Outstanding Research). He serves as an Associate Editor for Frontiers in Neuroscience and a Session Chair or a PC Member for a number of conferences. \end{IEEEbiography}

\begin{thebibliography}{10}
\providecommand{\url}[1]{#1}
\csname url@samestyle\endcsname
\providecommand{\newblock}{\relax}
\providecommand{\bibinfo}[2]{#2}
\providecommand{\BIBentrySTDinterwordspacing}{\spaceskip=0pt\relax}
\providecommand{\BIBentryALTinterwordstretchfactor}{4}
\providecommand{\BIBentryALTinterwordspacing}{\spaceskip=\fontdimen2\font plus
\BIBentryALTinterwordstretchfactor\fontdimen3\font minus \fontdimen4\font\relax}
\providecommand{\BIBforeignlanguage}[2]{{%
\expandafter\ifx\csname l@#1\endcsname\relax
\typeout{** WARNING: IEEEtran.bst: No hyphenation pattern has been}%
\typeout{** loaded for the language `#1'. Using the pattern for}%
\typeout{** the default language instead.}%
\else
\language=\csname l@#1\endcsname
\fi
#2}}
\providecommand{\BIBdecl}{\relax}
\BIBdecl

\bibitem{hu2022many}
P.~Hu, S.~Niklaus, S.~Sclaroff, and K.~Saenko, ``Many-to-many splatting for efficient video frame interpolation,'' in \emph{Proceedings of the IEEE/CVF Conference on Computer Vision and Pattern Recognition}, 2022, pp. 3553--3562.

\bibitem{huang2022real}
Z.~Huang, T.~Zhang, W.~Heng, B.~Shi, and S.~Zhou, ``Real-time intermediate flow estimation for video frame interpolation,'' in \emph{European Conference on Computer Vision}.\hskip 1em plus 0.5em minus 0.4em\relax Springer, 2022, pp. 624--642.

\bibitem{jiang2018super}
H.~Jiang, D.~Sun, V.~Jampani, M.-H. Yang, E.~Learned-Miller, and J.~Kautz, ``Super slomo: High quality estimation of multiple intermediate frames for video interpolation,'' in \emph{Proceedings of the IEEE conference on computer vision and pattern recognition}, 2018, pp. 9000--9008.

\bibitem{jin2023enhanced}
X.~Jin, L.~Wu, G.~Shen, Y.~Chen, J.~Chen, J.~Koo, and C.-h. Hahm, ``Enhanced bi-directional motion estimation for video frame interpolation,'' in \emph{Proceedings of the IEEE/CVF Winter Conference on Applications of Computer Vision}, 2023, pp. 5049--5057.

\bibitem{park2020bmbc}
J.~Park, K.~Ko, C.~Lee, and C.-S. Kim, ``Bmbc: Bilateral motion estimation with bilateral cost volume for video interpolation,'' in \emph{Computer Vision--ECCV 2020: 16th European Conference, Glasgow, UK, August 23--28, 2020, Proceedings, Part XIV 16}.\hskip 1em plus 0.5em minus 0.4em\relax Springer, 2020, pp. 109--125.

\bibitem{park2021asymmetric}
J.~Park, C.~Lee, and C.-S. Kim, ``Asymmetric bilateral motion estimation for video frame interpolation,'' in \emph{Proceedings of the IEEE/CVF international conference on computer vision}, 2021, pp. 14\,539--14\,548.

\bibitem{shen2024ladder}
T.~Shen, D.~Li, Z.~Gao, L.~Tian, and E.~Barsoum, ``Ladder: An efficient framework for video frame interpolation,'' \emph{arXiv preprint arXiv:2404.11108}, 2024.

\bibitem{niklaus2017video}
S.~Niklaus, L.~Mai, and F.~Liu, ``Video frame interpolation via adaptive convolution,'' in \emph{Proceedings of the IEEE conference on computer vision and pattern recognition}, 2017, pp. 670--679.

\bibitem{niklaus2021revisiting}
S.~Niklaus, L.~Mai, and O.~Wang, ``Revisiting adaptive convolutions for video frame interpolation,'' in \emph{Proceedings of the IEEE/CVF winter conference on applications of computer vision}, 2021, pp. 1099--1109.

\bibitem{amir2017low}
A.~Amir, B.~Taba, D.~Berg, T.~Melano, J.~McKinstry, C.~Di~Nolfo, T.~Nayak, A.~Andreopoulos, G.~Garreau, M.~Mendoza \emph{et~al.}, ``A low power, fully event-based gesture recognition system,'' in \emph{Proceedings of the IEEE conference on computer vision and pattern recognition}, 2017, pp. 7243--7252.

\bibitem{tulyakov2021time}
S.~Tulyakov, D.~Gehrig, S.~Georgoulis, J.~Erbach, M.~Gehrig, Y.~Li, and D.~Scaramuzza, ``Time lens: Event-based video frame interpolation,'' in \emph{Proceedings of the IEEE/CVF conference on computer vision and pattern recognition}, 2021, pp. 16\,155--16\,164.

\bibitem{tulyakov2022time}
S.~Tulyakov, A.~Bochicchio, D.~Gehrig, S.~Georgoulis, Y.~Li, and D.~Scaramuzza, ``Time lens++: Event-based frame interpolation with parametric non-linear flow and multi-scale fusion,'' in \emph{Proceedings of the IEEE/CVF Conference on Computer Vision and Pattern Recognition}, 2022, pp. 17\,755--17\,764.

\bibitem{kim2023event}
T.~Kim, Y.~Chae, H.-K. Jang, and K.-J. Yoon, ``Event-based video frame interpolation with cross-modal asymmetric bidirectional motion fields,'' in \emph{Proceedings of the IEEE/CVF Conference on Computer Vision and Pattern Recognition}, 2023, pp. 18\,032--18\,042.

\bibitem{ma2024timelens}
Y.~Ma, S.~Guo, Y.~Chen, T.~Xue, and J.~Gu, ``Timelens-xl: Real-time event-based video frame interpolation with large motion,'' in \emph{European Conference on Computer Vision}.\hskip 1em plus 0.5em minus 0.4em\relax Springer, 2024, pp. 178--194.

\bibitem{liu2024video}
Y.~Liu, Y.~Deng, H.~Chen, and Z.~Yang, ``Video frame interpolation via direct synthesis with the event-based reference,'' in \emph{Proceedings of the IEEE/CVF Conference on Computer Vision and Pattern Recognition}, 2024, pp. 8477--8487.

\bibitem{song2020denoising}
J.~Song, C.~Meng, and S.~Ermon, ``Denoising diffusion implicit models,'' \emph{arXiv preprint arXiv:2010.02502}, 2020.

\bibitem{song2020score}
Y.~Song, J.~Sohl-Dickstein, D.~P. Kingma, A.~Kumar, S.~Ermon, and B.~Poole, ``Score-based generative modeling through stochastic differential equations,'' \emph{arXiv preprint arXiv:2011.13456}, 2020.

\bibitem{ho2022video}
J.~Ho, T.~Salimans, A.~Gritsenko, W.~Chan, M.~Norouzi, and D.~J. Fleet, ``Video diffusion models,'' \emph{Advances in Neural Information Processing Systems}, vol.~35, pp. 8633--8646, 2022.

\bibitem{singer2022make}
U.~Singer, A.~Polyak, T.~Hayes, X.~Yin, J.~An, S.~Zhang, Q.~Hu, H.~Yang, O.~Ashual, O.~Gafni \emph{et~al.}, ``Make-a-video: Text-to-video generation without text-video data,'' \emph{arXiv preprint arXiv:2209.14792}, 2022.

\bibitem{yin2023nuwa}
S.~Yin, C.~Wu, H.~Yang, J.~Wang, X.~Wang, M.~Ni, Z.~Yang, L.~Li, S.~Liu, F.~Yang \emph{et~al.}, ``Nuwa-xl: Diffusion over diffusion for extremely long video generation,'' \emph{arXiv preprint arXiv:2303.12346}, 2023.

\bibitem{xing2024dynamicrafter}
J.~Xing, M.~Xia, Y.~Zhang, H.~Chen, W.~Yu, H.~Liu, G.~Liu, X.~Wang, Y.~Shan, and T.-T. Wong, ``Dynamicrafter: Animating open-domain images with video diffusion priors,'' in \emph{European Conference on Computer Vision}.\hskip 1em plus 0.5em minus 0.4em\relax Springer, 2024, pp. 399--417.

\bibitem{voleti2022mcvd}
V.~Voleti, A.~Jolicoeur-Martineau, and C.~Pal, ``Mcvd-masked conditional video diffusion for prediction, generation, and interpolation,'' \emph{Advances in neural information processing systems}, vol.~35, pp. 23\,371--23\,385, 2022.

\bibitem{danier2024ldmvfi}
D.~Danier, F.~Zhang, and D.~Bull, ``Ldmvfi: Video frame interpolation with latent diffusion models,'' in \emph{Proceedings of the AAAI Conference on Artificial Intelligence}, vol.~38, no.~2, 2024, pp. 1472--1480.

\bibitem{lyu2024frame}
Z.~Lyu, M.~Li, J.~Jiao, and C.~Chen, ``Frame interpolation with consecutive brownian bridge diffusion,'' in \emph{Proceedings of the 32nd ACM International Conference on Multimedia}, 2024, pp. 3449--3458.

\bibitem{cheng2020video}
X.~Cheng and Z.~Chen, ``Video frame interpolation via deformable separable convolution,'' in \emph{Proceedings of the AAAI Conference on Artificial Intelligence}, vol.~34, no.~07, 2020, pp. 10\,607--10\,614.

\bibitem{cheng2021multiple}
------, ``Multiple video frame interpolation via enhanced deformable separable convolution,'' \emph{IEEE Transactions on Pattern Analysis and Machine Intelligence}, vol.~44, no.~10, pp. 7029--7045, 2021.

\bibitem{chen2021pdwn}
Z.~Chen, R.~Wang, H.~Liu, and Y.~Wang, ``Pdwn: Pyramid deformable warping network for video interpolation,'' \emph{IEEE Open Journal of Signal Processing}, vol.~2, pp. 413--424, 2021.

\bibitem{bao2019memc}
W.~Bao, W.-S. Lai, X.~Zhang, Z.~Gao, and M.-H. Yang, ``Memc-net: Motion estimation and motion compensation driven neural network for video interpolation and enhancement,'' \emph{IEEE transactions on pattern analysis and machine intelligence}, vol.~43, no.~3, pp. 933--948, 2019.

\bibitem{danier2022st}
D.~Danier, F.~Zhang, and D.~Bull, ``St-mfnet: A spatio-temporal multi-flow network for frame interpolation,'' in \emph{Proceedings of the IEEE/CVF Conference on Computer Vision and Pattern Recognition}, 2022, pp. 3521--3531.

\bibitem{lu2022video}
L.~Lu, R.~Wu, H.~Lin, J.~Lu, and J.~Jia, ``Video frame interpolation with transformer,'' in \emph{Proceedings of the IEEE/CVF Conference on Computer Vision and Pattern Recognition}, 2022, pp. 3532--3542.

\bibitem{zhang2023extracting}
G.~Zhang, Y.~Zhu, H.~Wang, Y.~Chen, G.~Wu, and L.~Wang, ``Extracting motion and appearance via inter-frame attention for efficient video frame interpolation,'' in \emph{Proceedings of the IEEE/CVF Conference on Computer Vision and Pattern Recognition}, 2023, pp. 5682--5692.

\bibitem{li2023amt}
Z.~Li, Z.-L. Zhu, L.-H. Han, Q.~Hou, C.-L. Guo, and M.-M. Cheng, ``Amt: All-pairs multi-field transforms for efficient frame interpolation,'' in \emph{Proceedings of the IEEE/CVF Conference on Computer Vision and Pattern Recognition}, 2023, pp. 9801--9810.

\bibitem{plack2023frame}
M.~Plack, K.~M. Briedis, A.~Djelouah, M.~B. Hullin, M.~Gross, and C.~Schroers, ``Frame interpolation transformer and uncertainty guidance,'' in \emph{Proceedings of the IEEE/CVF Conference on Computer Vision and Pattern Recognition}, 2023, pp. 9811--9821.

\bibitem{paikin2021efi}
G.~Paikin, Y.~Ater, R.~Shaul, and E.~Soloveichik, ``Efi-net: Video frame interpolation from fusion of events and frames,'' in \emph{Proceedings of the IEEE/CVF conference on computer vision and pattern recognition}, 2021, pp. 1291--1301.

\bibitem{gao2022superfast}
Y.~Gao, S.~Li, Y.~Li, Y.~Guo, and Q.~Dai, ``Superfast: 200$\times$ video frame interpolation via event camera,'' \emph{IEEE transactions on pattern analysis and machine intelligence}, vol.~45, no.~6, pp. 7764--7780, 2022.

\bibitem{rombach2022high}
R.~Rombach, A.~Blattmann, D.~Lorenz, P.~Esser, and B.~Ommer, ``High-resolution image synthesis with latent diffusion models,'' in \emph{Proceedings of the IEEE/CVF conference on computer vision and pattern recognition}, 2022, pp. 10\,684--10\,695.

\bibitem{jain2024video}
S.~Jain, D.~Watson, E.~Tabellion, B.~Poole, J.~Kontkanen \emph{et~al.}, ``Video interpolation with diffusion models,'' in \emph{Proceedings of the IEEE/CVF Conference on Computer Vision and Pattern Recognition}, 2024, pp. 7341--7351.

\bibitem{wang2024framer}
W.~Wang, Q.~Wang, K.~Zheng, H.~Ouyang, Z.~Chen, B.~Gong, H.~Chen, Y.~Shen, and C.~Shen, ``Framer: Interactive frame interpolation,'' \emph{arXiv preprint arXiv:2410.18978}, 2024.

\bibitem{shen2024dreammover}
L.~Shen, T.~Liu, H.~Sun, X.~Ye, B.~Li, J.~Zhang, and Z.~Cao, ``Dreammover: Leveraging the prior of diffusion models for image interpolation with large motion,'' in \emph{European Conference on Computer Vision}.\hskip 1em plus 0.5em minus 0.4em\relax Springer, 2024, pp. 336--353.

\bibitem{huang2024motion}
Z.~Huang, Y.~Yu, L.~Yang, C.~Qin, B.~Zheng, X.~Zheng, Z.~Zhou, Y.~Wang, and W.~Yang, ``Motion-aware latent diffusion models for video frame interpolation,'' in \emph{Proceedings of the 32nd ACM International Conference on Multimedia}, 2024, pp. 1043--1052.

\bibitem{howard2017mobilenets}
A.~G. Howard, M.~Zhu, B.~Chen, D.~Kalenichenko, W.~Wang, T.~Weyand, M.~Andreetto, and H.~Adam, ``Mobilenets: Efficient convolutional neural networks for mobile vision applications,'' \emph{arXiv preprint arXiv:1704.04861}, 2017.

\bibitem{tu2022maxvit}
Z.~Tu, H.~Talebi, H.~Zhang, F.~Yang, P.~Milanfar, A.~Bovik, and Y.~Li, ``Maxvit: Multi-axis vision transformer,'' in \emph{European conference on computer vision}.\hskip 1em plus 0.5em minus 0.4em\relax Springer, 2022, pp. 459--479.

\bibitem{van2017neural}
A.~Van Den~Oord, O.~Vinyals \emph{et~al.}, ``Neural discrete representation learning,'' \emph{Advances in neural information processing systems}, vol.~30, 2017.

\bibitem{zhang2018unreasonable}
R.~Zhang, P.~Isola, A.~A. Efros, E.~Shechtman, and O.~Wang, ``The unreasonable effectiveness of deep features as a perceptual metric,'' in \emph{Proceedings of the IEEE conference on computer vision and pattern recognition}, 2018, pp. 586--595.

\bibitem{isola2017image}
P.~Isola, J.-Y. Zhu, T.~Zhou, and A.~A. Efros, ``Image-to-image translation with conditional adversarial networks,'' in \emph{Proceedings of the IEEE conference on computer vision and pattern recognition}, 2017, pp. 1125--1134.

\bibitem{xia2024diffi2i}
B.~Xia, Y.~Zhang, S.~Wang, Y.~Wang, X.~Wu, Y.~Tian, W.~Yang, R.~Timotfe, and L.~Van~Gool, ``Diffi2i: efficient diffusion model for image-to-image translation,'' \emph{IEEE Transactions on Pattern Analysis and Machine Intelligence}, 2024.

\bibitem{xue2019video}
T.~Xue, B.~Chen, J.~Wu, D.~Wei, and W.~T. Freeman, ``Video enhancement with task-oriented flow,'' \emph{International Journal of Computer Vision}, vol. 127, pp. 1106--1125, 2019.

\bibitem{nah2017deep}
S.~Nah, T.~Hyun~Kim, and K.~Mu~Lee, ``Deep multi-scale convolutional neural network for dynamic scene deblurring,'' in \emph{Proceedings of the IEEE conference on computer vision and pattern recognition}, 2017, pp. 3883--3891.

\bibitem{choi2020channel}
M.~Choi, H.~Kim, B.~Han, N.~Xu, and K.~M. Lee, ``Channel attention is all you need for video frame interpolation,'' in \emph{Proceedings of the AAAI conference on artificial intelligence}, vol.~34, no.~07, 2020, pp. 10\,663--10\,671.

\bibitem{gehrig2020video}
D.~Gehrig, M.~Gehrig, J.~Hidalgo-Carri{\'o}, and D.~Scaramuzza, ``Video to events: Recycling video datasets for event cameras,'' in \emph{Proceedings of the IEEE/CVF Conference on Computer Vision and Pattern Recognition}, 2020, pp. 3586--3595.

\bibitem{stoffregen2020reducing}
T.~Stoffregen, C.~Scheerlinck, D.~Scaramuzza, T.~Drummond, N.~Barnes, L.~Kleeman, and R.~Mahony, ``Reducing the sim-to-real gap for event cameras,'' in \emph{Computer Vision--ECCV 2020: 16th European Conference, Glasgow, UK, August 23--28, 2020, Proceedings, Part XXVII 16}.\hskip 1em plus 0.5em minus 0.4em\relax Springer, 2020, pp. 534--549.

\bibitem{kong2022ifrnet}
L.~Kong, B.~Jiang, D.~Luo, W.~Chu, X.~Huang, Y.~Tai, C.~Wang, and J.~Yang, ``Ifrnet: Intermediate feature refine network for efficient frame interpolation,'' in \emph{Proceedings of the IEEE/CVF Conference on Computer Vision and Pattern Recognition}, 2022, pp. 1969--1978.

\bibitem{jin2023unified}
X.~Jin, L.~Wu, J.~Chen, Y.~Chen, J.~Koo, and C.-h. Hahm, ``A unified pyramid recurrent network for video frame interpolation,'' in \emph{Proceedings of the IEEE/CVF Conference on Computer Vision and Pattern Recognition}, 2023, pp. 1578--1587.

\bibitem{wu2022video}
S.~Wu, K.~You, W.~He, C.~Yang, Y.~Tian, Y.~Wang, Z.~Zhang, and J.~Liao, ``Video interpolation by event-driven anisotropic adjustment of optical flow,'' in \emph{European Conference on Computer Vision}.\hskip 1em plus 0.5em minus 0.4em\relax Springer, 2022, pp. 267--283.

\bibitem{sun2022event}
L.~Sun, C.~Sakaridis, J.~Liang, Q.~Jiang, K.~Yang, P.~Sun, Y.~Ye, K.~Wang, and L.~V. Gool, ``\textrm{Event-based fusion for motion deblurring with cross-modal attention},'' in \emph{European conference on computer vision}.\hskip 1em plus 0.5em minus 0.4em\relax Springer, 2022, pp. 412--428.

\bibitem{tao2018scale}
X.~Tao, H.~Gao, X.~Shen, J.~Wang, and J.~Jia, ``\textrm{Scale-recurrent network for deep image deblurring},'' in \emph{Proceedings of the IEEE conference on computer vision and pattern recognition}, 2018, pp. 8174--8182.

\bibitem{chen2021hinet}
L.~Chen, X.~Lu, J.~Zhang, X.~Chu, and C.~Chen, ``\textrm{Hinet: Half instance normalization network for image restoration},'' in \emph{Proceedings of the IEEE/CVF conference on computer vision and pattern recognition}, 2021, pp. 182--192.

\bibitem{li2022learning}
D.~Li, Y.~Zhang, K.~C. Cheung, X.~Wang, H.~Qin, and H.~Li, ``\textrm{Learning degradation representations for image deblurring},'' in \emph{European conference on computer vision}.\hskip 1em plus 0.5em minus 0.4em\relax Springer, 2022, pp. 736--753.

\bibitem{chen2022simple}
L.~Chen, X.~Chu, X.~Zhang, and J.~Sun, ``\textrm{Simple baselines for image restoration},'' in \emph{European conference on computer vision}.\hskip 1em plus 0.5em minus 0.4em\relax Springer, 2022, pp. 17--33.

\bibitem{fang2023self}
Z.~Fang, F.~Wu, W.~Dong, X.~Li, J.~Wu, and G.~Shi, ``\textrm{Self-supervised non-uniform kernel estimation with flow-based motion prior for blind image deblurring},'' in \emph{Proceedings of the IEEE/CVF conference on computer vision and pattern recognition}, 2023, pp. 18\,105--18\,114.

\bibitem{shang2021bringing}
W.~Shang, D.~Ren, D.~Zou, J.~S. Ren, P.~Luo, and W.~Zuo, ``\textrm{Bringing events into video deblurring with non-consecutively blurry frames},'' in \emph{Proceedings of the IEEE/CVF international conference on computer vision}, 2021, pp. 4531--4540.

\bibitem{haoyu2020learning}
C.~Haoyu, T.~Minggui, S.~Boxin, W.~YIzhou, and H.~Tiejun, ``\textrm{Learning to deblur and generate high frame rate video with an event camera},'' \emph{arXiv preprint arXiv:2003.00847}, 2020.

\bibitem{sun2024motion}
Z.~Sun, X.~Fu, L.~Huang, A.~Liu, and Z.-J. Zha, ``\textrm{Motion Aware Event Representation-Driven Image Deblurring},'' in \emph{European Conference on Computer Vision}.\hskip 1em plus 0.5em minus 0.4em\relax Springer, 2024, pp. 418--435.

\bibitem{yang2024motion}
W.~Yang, J.~Wu, J.~Ma, L.~Li, and G.~Shi, ``\textrm{Motion deblurring via spatial-temporal collaboration of frames and events},'' in \emph{Proceedings of the AAAI Conference on Artificial Intelligence}, vol.~38, no.~7, 2024, pp. 6531--6539.

\bibitem{xia2023diffir}
B.~Xia, Y.~Zhang, S.~Wang, Y.~Wang, X.~Wu, Y.~Tian, W.~Yang, and L.~Van~Gool, ``\textrm{Diffir: Efficient diffusion model for image restoration},'' in \emph{Proceedings of the IEEE/CVF International Conference on Computer Vision}, 2023, pp. 13\,095--13\,105.

\bibitem{ren2023multiscale}
M.~Ren, M.~Delbracio, H.~Talebi, G.~Gerig, and P.~Milanfar, ``\textrm{Multiscale structure guided diffusion for image deblurring},'' in \emph{Proceedings of the IEEE/CVF International Conference on Computer Vision}, 2023, pp. 10\,721--10\,733.

\bibitem{chen2024hierarchical}
Z.~Chen, Y.~Zhang, D.~Liu, J.~Gu, L.~Kong, X.~Yuan \emph{et~al.}, ``\textrm{Hierarchical integration diffusion model for realistic image deblurring},'' \emph{Advances in neural information processing systems}, vol.~36, 2024.

\bibitem{chen2024efficient}
K.~Chen and Y.~Liu, ``\textrm{Efficient image deblurring networks based on diffusion models},'' \emph{arXiv preprint arXiv:2401.05907}, 2024.

\end{thebibliography}
\end{document}